\documentclass[runningheads]{llncs}
\usepackage{graphicx}

\usepackage{tikz}
\usepackage{comment}
\usepackage{amsmath,amssymb, bm}
\usepackage{color}

\usepackage[width=122mm,left=12mm,paperwidth=146mm,height=193mm,top=12mm,paperheight=217mm]{geometry}
\usepackage{times}
\usepackage{epsfig}
\usepackage{graphicx}

\usepackage{filecontents,lipsum}
\usepackage{comment}
\usepackage{amssymb}
\usepackage{amsmath}
\usepackage{mathtools}
\usepackage{breqn}
\usepackage{varwidth}

\usepackage{array}
\usepackage{multirow}
\usepackage{footnote}
\usepackage{cases}
\usepackage{listings}
\usepackage{booktabs}
\usepackage{tabularx}
\usepackage{breqn}
\usepackage{float}
\usepackage{tabularx}
\usepackage{enumerate}

\usepackage{afterpage}
\usepackage{lipsum}
\usepackage{capt-of}
\usepackage{scrextend}
\usepackage{mathrsfs}

\usepackage{epsfig}	
\usepackage{booktabs}
\usepackage[belowskip=-1pt,aboveskip=0pt]{caption}
\usepackage{subcaption}
\usepackage{breqn}
\usepackage{array}
\usepackage{xcolor}
\usepackage{color, colortbl}
\usepackage[linesnumbered,ruled,vlined]{algorithm2e}
\usepackage{algcompatible}
\usepackage[ruled,vlined]{algorithm2e}
\usepackage{algpseudocode}
\usepackage[figuresright]{rotating}	

\newcolumntype{P}[1]{>{\centering\arraybackslash}p{#1}}
\newcolumntype{M}[1]{>{\centering\arraybackslash}m{#1}}

\DeclareGraphicsExtensions{.pdf,.jpeg,.png,.tif}
\makeatletter
\def\BState{\State\hskip-\ALG@thistlm}

\SetCommentSty{mycommfont}
\newcommand{\mycc}{\cellcolor{lightgray}}
\newcommand{\myrr}{\cellcolor{red!15}}
\newcommand{\mybb}{\cellcolor{blue!15}}

\definecolor{Gray}{gray}{0.4}
\usepackage[utf8]{inputenc}
\usepackage[english]{babel}
\usepackage{framed,multirow}
\usepackage{amssymb}
\usepackage{latexsym}
\usepackage{breqn}
\usepackage[utf8]{inputenc}
\usepackage[english]{babel}
\usepackage{url}
\usepackage{xcolor}
\definecolor{newcolor}{rgb}{.8,.349,.1}

\begin{document}
	
	\pagestyle{headings}
	\mainmatter

	\title{Physics Informed and Data Driven Simulation of Underwater Images via Residual Learning}
	
	\author{Tanmoy Mondal\inst{1} \and Ricardo Mendoza\inst{2} \and Lucas Drumetz\inst{1}}
	\institute{TOMS Team, IMT Atlantique, Brest, France, \\
	\email{tanmoy.besu@gmail.com, lucas.drumetz@imt-atlantique.fr}
	\and
	Cervval, Brest, France, \\
	\email{mendoza@cervval.com}
	}
	
	
	\maketitle
	\begin{abstract}
		In general, underwater images suffer from color distortion and low contrast, because light is attenuated and backscattered as it propagates through water (differently depending on wavelength and on the properties of the water body).  
		An existing simple degradation model (similar to atmospheric image ``hazing'' effects), though helpful, is not sufficient to properly represent the underwater image degradation because there are unaccounted for and non-measurable factors e.g. scattering of light due to turbidity of water, reflective characteristics of turbid medium etc.
		We propose a deep learning-based architecture to automatically simulate the underwater effects where only a dehazing-like image formation equation is known to the network, and the additional degradation due to the other unknown factors if inferred in a data-driven way. We only use RGB images (because in real-time scenario depth image is not available) to estimate the depth image.
		
		
		For testing, we have proposed (due to the lack of real underwater image datasets) a complex image formation model/equation to manually generate images that resemble real underwater images (used as ground truth). However, only the classical image formation equation (the one used for image dehazing) is informed to the network. This mimics the fact that in a real scenario, the physics are never completely known and only simplified models are known.
		Thanks to the ground truth, generated by a complex image formation equation, we could successfully perform a qualitative and quantitative evaluation of proposed technique, compared to other purely data driven approaches. For code and dataset, see: {\color{blue}\url{https://github.com/anoynymREVIEW/underwater_simulation.git}}
		
	\end{abstract}
	
	\section{Introduction}
	Underwater images lack contrast and contain a different color palette from usual natural images. This occurs because the light spectrum is selectively absorbed and scattered (mainly due to the floating particles in the water) during the propagation of light in water. 
	The attenuation of light highly depends on the wavelength, which varies with respect to the water type, depth and the distance which light has to travel to illuminate the object \cite{Mobley2004}. Underwater images also depend on the $3D$ structure of the scene and floating particles in the water, which makes it very difficult to model the underwater scattering phenomena. 
	The wavelength-dependent attenuation of light causes color distortions and are directly related to the objects' distances from the source of light. Furthermore, light scattering introduces an additional factor into the image which inherently decreases the image's contrast. Light scattering is directly related to the object's distance from the light's source, which explains why these phenomena cannot be easily globally corrected \cite{Akkaynak2017}. Moreover, the attenuation parameters of the water medium are highly affected by seasonal, geographical and climate variations. These variations of attenuation parameters were categorized into $10$ categories by Jerlov \cite{Jerlov1976}. \\
	Furthermore, underwater image formation also gets affected by the scattering of light due to reflective characteristics of the turbid medium, among others. More importantly, these factors are not measurable and difficult to incorporate within a mathematical model. 
	In this work, we propose an end-to-end deep learning-based physics-informed model to simulate the underwater effects. A classical image formation equation is hard coded into the network, and it estimates additional non-measurable and complex factors which influence the underwater image degradation. We have only used clean \emph{RGB} atmospheric images and have estimated the depth image (the depth values of each image pixel) from it. Because in a real scenario, we do not have the access to each pixel's depth values (because we need to use \emph{RGB-D} cameras, e.g. Microsoft Kinect, since classical \emph{RGB-D} cameras cannot be used for underwater imaging). The estimated depth image is fed to the physical model part of the network.\\
	Due to the inherent difficulty of obtaining pairs of real-world clean/degraded images in an underwater context, we propose in addition a complex image formation model/equation to manually generate images that resemble real-world underwater images. The generated images are used as ground truth for our experiments. 
	This work is a proof of concept, where the objective is to simulate images as close as possible to the observed data (which we manually generate using our proposed physical model, mentioned in Equation~\eqref{eq_3_4}), while capturing unaccounted and unmeasurable physical effects in a data-driven manner. 
	A rich dataset of clean-degraded image pairs is created to train a neural network model that will be used as a simulator to generate varied underwater images, parameterized by a few user given parameters.\\
	This way, once trained, our model can be effectively generate realistic rare underwater images and provide an efficient physically explainable emulator.
	To the best of our knowledge, there are no other research works on the simulation of underwater images. The proposed physics-data-driven method to simulate under water image degradation effects using a deep neural network is a novel technique. The contribution of this work are as follows: \emph{i)} We propose a complex image formation model/equation to manually generate images that resemble real underwater images (see section~\ref{sec:image_formation}) which is used as ground truth. \emph{ii)} Then we propose a deep neural network based architecture to simulate the complex under water imaging effect by informing the network about the classical image degradation model (see section~\ref{proposed_method}), to make it interpretable and able to capture missing degradation in a data-driven way. We have further analyzed the influence and effectiveness of each block on the overall performance of the network, and compared it to simulators obtained from a number of other data-driven deep learning models. 

	\section{Related Work}
	In this section, we discuss the literature related to the correction of haze-related degradation of terrestrial images, which has a thin connection with underwater image degradations. 
	In case of terrestrial images, in the presence of fog, haze or turbulence, the transmitting light gets diffracted and scattered while passing through the atmosphere \cite{Schwartzman2017}. Many image dehazing techniques in the literature \cite{Berman2016}, \cite{He2011b}, \cite{Kaiming2009}, \cite{Fattal2014}, \cite{Nishino2012}, \cite{Tan2008} take as input a single hazy image and estimate the unknown distance map, from which the clear image of the scene is generated. Researchers have proposed several priors in order to solve the ill-posed inverse problem. It is assumed that transmission is color independent for terrestrial images but that is not the case for underwater images. One of such priors is the so-called ``Dark Channel Prior (DCP)'' \cite{He2011b} \cite{Kaiming2009} which assumes that within small image patches, at least one pixel has a low value in some color channel. This minimal value is used to estimate the distance. \\
	This concept has been widely used in the processing of underwater images \cite{Carlevaris-Bianco2010}, \cite{Chiang2012}, \cite{Drews-Jr2013}, \cite{Galdran2015}, \cite{Lu2015}. 
	Although, it works well in case of terrestrial images, the same underlying assumption does not hold in many underwater scenarios. For example, a bright sand foreground has high values in all color channels and is often mistaken to have low transmission despite being close to the camera. 
	Because of these reasons, 
	\cite{Peng2017} proposed a depth estimation method for underwater scenes which is based on image blurriness and light absorption. Although such prior is physically valid, it has limited efficiency in texture-less areas. There are a few more works in the literature which focus on perceptually pleasing results e.g. \cite{Ancuti2012}, \cite{Lu2013a}, \cite{Ancuti2016} but have not shown color consistency which is required for scientific measurements. 
	There have been only few number of attempts of using deep neural networks \cite{Cai2016}, \cite{Li2017a}, \cite{Chen2019a}, \cite{Ren2016} for the restoration of terrestrial haze images and even less for underwater images. In \cite{Shin2016}, authors estimate the ambient light and transmission in underwater images by using a classical convolution neural network (CNN) architecture which is further used to dehaze underwater images. \\
	The simulation of underwater images has strong resemblance with the task of \emph{Image-to-Image (I2I)} translation which aims to learn a mapping between different visual domains. This task is challenging for two main reasons. First, it is either difficult to collect aligned training image pairs (e.g. day scene $\leftrightarrow$ night scene), or they simply do not exist (e.g. artwork $\leftrightarrow$ real photo). Secondly, many such mappings are inherently multi-modal i.e. a single input may correspond to multiple possible outputs. Several techniques exists in the literature to address these issues. The \emph{Pix2Pix} architecture \cite{Isola2016} applies conditional generative adversarial networks to \emph{I2I} translation problems by using paired image data. There are a number of recent works \cite{Choi2018}, \cite{Taigman2017}, \cite{Liu2017}, \cite{Yi2017}, \cite{Isola2016} which are based on the paired training data for learning \emph{I2I} translation. These techniques produces a single output, conditioned on the given input image. To train with unpaired data, CycleGAN \cite{Zhu2017a}, DiscoGAN \cite{Kim2017} leverage cycle consistency to regularize the training. Another set of unpaired I2I translation techniques either generate one (e.g. UNIT \cite{Liu2017}) or many output images (e.g. MUNIT \cite{Huang2018b} and DRIT \cite{Lee2018a}) from a given input images, also by leveraging cycle consistency to regularize the training. \\
	On the other hand, BicycleGAN \cite{Zhu2017b} (only applicable to problems with paired training data) enforces a bijection mapping between the latent and target space to tackle the mode collapse problem. Contrary to all these approaches in the literature, the proposed neural network architecture incorporates a physics-informed classical image degradation model to drive the network and helps him simulate more complex image degradation model whose formulation and related parameters are unknown.   
	
	\section{Image Formation Model}
	\label{sec:image_formation}
	\subsection{Classical Simulation Model}
	In this section, we describe the widely used atmospheric image formation model developed in \cite{Schechner2005}. For each color channel $c \in \{R,G,B\}$, the image intensity at each pixel is composed of two components: the attenuated signal and the veiling light. 
	\begin{equation}
	I_c(\textbf{x}) = t_c(\textbf{x})J_c(\textbf{x})+(1-t_c(\textbf{x})).A_c
	\label{eq_1}
	\end{equation}
	where bold letters denotes vectors, $\textbf{x}$ is the pixel coordinate, $I_c$ is the acquired image value in the color channel $c$, $t_c$ is the transmission of the color channel, and $J_c$ is the object radiance or clean image. The global veiling-light/atmospheric light component $A_c$ is the scene value in the areas with no objects $(t_c=0, \forall c \in \{R, G, B\})$. The transmission depends on the object's distance $z(\textbf{x})$ and the attenuation coefficient of the medium for each channel i.e. $\beta_c$:
	\begin{equation}
	t_c(\textbf{x}) = \exp^{-\beta_c z(\textbf{x})}
	\label{eq_2}
	\end{equation}
	In the ocean, the attenuation of red colors can be an order of magnitude larger than the attenuation of blue and green \cite{Mobley2016}. Hence, contrary to the common assumption in single image dehazing, the transmission $\textbf{t} = (t_R, t_G, t_B)$ is wavelength dependent.
	
	The attenuation of light in underwater is not constant and varies with the change in geography, seasons and climate. The attenuation coefficient ($\beta$) is dependent on wavelength of various water types.
	Based on the water clarity, Jerlov \cite{Jerlov1976a} proposed a classification scheme for oceanic waters where open ocean waters are classified into class \textbf{I}, \textbf{IA}, \textbf{IB}, \textbf{II} and \textbf{III}. He also defined the water type $\textbf{1}$ through $\textbf{9}$ for coastal waters. Type \textbf{I} is the clearest and type \textbf{III} is the most turbid open ocean water. Similarly, for coastal water, type $\textbf{1}$ is the clearest and type $\textbf{9}$ is the most turbid. We use three attenuation coefficients: i.e. $\beta_R, \beta_G, \beta_B$ corresponding to $R$$G$$B$ channels for our work.

	\subsection{Underwater Image Simulation Model}
\begin{figure}[!h]
		\centering{\includegraphics[trim = 0.0cm 0.0cm 0.0cm 0.0cm, clip, scale = 0.25]{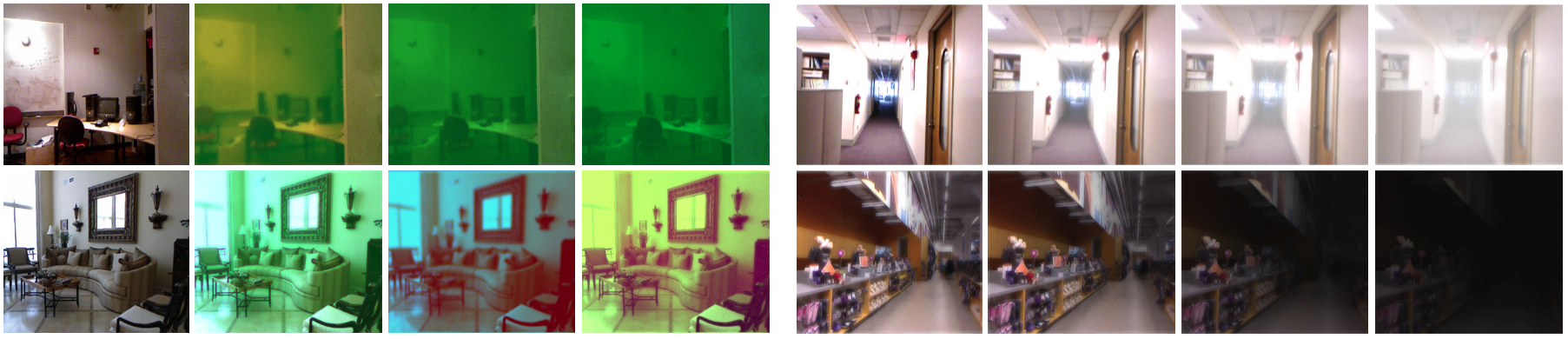}}
		\centering	
		\caption{
			Examples of degradations that can be simulated using our approach on NYU Depth v2 dataset. Left: First row - underwater degradation (with increased water attenuation); Second row - diverse attenuation-lighting-scattering configurations. Right: first row - scatter-enriched fog, second row - smoke-like degraded environments.
		}
		\label{fig:Fig_3-3_1}
	\end{figure}

	Obtaining a dataset of underwater images along with the ground truth information for depth, composition of veiling light and transmission coefficients is a challenging and expensive task. Here we describe our contribution to simulate underwater images, which is used to generate the ground truth of our underwater image dataset. 
	This dataset aims to approximate certain complex underwater phenomena such as forward photon scattering in an absorbing medium \cite{fu2001mie} and turbidity in water due to colored dissolved matter \cite{bricaud1998variations}, which inherently cause complementary image degradation (Fig.~\ref{fig:Fig_3-3_1}).
	
	Our proposed model for underwater image simulation initiates from the classic atmospheric image formation model (see Equation 1) and introduces the effect of forward scattering by replacing the known radiance (clear image) with the sum of two contributions: one due to direct or straight-path scene radiance $J'_c(\textbf{x}) = (1-k_c(\textbf{x})) J_c(\textbf{x})$ and another due to scattered light from scene radiance $J_c^{sct}(\textbf{x})$, where $k_c(\textbf{x})= \exp^{-\alpha_c z(\textbf{x})}$ modulates the depth-dependent likelihood that photons from the scene follow a straight-line path. $\alpha_c$ parameterizes scattering media (in an analogous fashion to a ``particle density''); we assume that all unscattered photons follow straight-line paths which are perpendicular to the degraded image plane (as is implicitly assumed in Equation~\ref{eq_1}).
	For forward scattering radiance, we follow an isotropic, non-polarized, Lambertian-inspired diffusion \& reflectance \cite{nayar1991surface} light transmission model. In addition, we see scattering interactions as occurring at the same depth as the radiance source. 
	In order to model how scattered photons travel from a single radiance source, we use a diffusive (bi-variate isotropic Gaussian) approximation to compute the likelihood that a scattered photon departing from the radiance source (clear image pixel coordinates) $\textbf{x}'=(x_1',x_2')$ with depth $z(\textbf{x}')$ arrives at pixel coordinates $\textbf{x}=(x_1,x_2)$ of the degraded image plane:
	\begin{equation}
	G_c(\textbf{x},\textbf{x}') = \frac{1}{ 2\pi(\gamma_c z(\textbf{x}') )} \exp{\left( -\frac{(x_1'-x_1)^2 + (x_2'-x_2)^2}{2(\gamma_c z(\textbf{x}'))^2} \right)},
	\label{eq_3_1}
	\end{equation}
	where $\gamma_c$ modulates the scattering relationship with distance for channel $c$ (acting as a proxy for scattering trajectory variance). Eq. 3 can be interpreted as the two dimensional distribution of the forward-scattered arriving intensity over the degraded image plane from a single radiance source; the larger the "on-plane" distance separating source pixel $\textbf{x}'$ from the recovery pixel $\textbf{x}$ is, the lower this likelihood becomes (see \cite{berrocal2007laser} for a more elaborate version approximating the Lorentz-Mie scattering phase function distribution on individual particles).  
	Furthermore, we view each clear image pixel $\textbf{x}'\in S$ as an individual radiance source, being each one subject to scattering dynamics independently (this holds if wave-like and quantum interactions are excluded). Therefore, for a given degraded image pixel $\textbf{x}$ and channel $c$, integrating the radiance contributions onto $\textbf{x}$ (e.i. the likelihood of a photon being scattered, multiplied by the likelihood of a scattered photon arriving at $\textbf{x}$, multiplied by the source radiance) from each source on $S$, provides an expectation of the recovered scattered signal:
	\begin{equation}
	J_c^{sct}(\textbf{x}) = \int_{\textbf{x}' \in S} k_c(\textbf{x}') J_c(\textbf{x}') G_c(\textbf{x},\textbf{x}')d\textbf{x}'.
	\label{eq_3_2}
	\end{equation}
	This way, the signal collected at degraded image pixel $\textbf{x}$ after considering signal attenuation, scattering effects and ambiance lighting is approximated by
	\begin{equation}
	I^{sct}_c(\textbf{x})= 
	( J_c^{sct}(\textbf{x}) + (1-k_c(\textbf{x})) J_c(\textbf{x}) ) t_c(\textbf{x}) + (1-t_c(\textbf{x}))A_c.
	\label{eq_3_3}
	\end{equation}
	Finally, we increase the turbid visual appearance of the degraded image by weighting $I^{sct}_c$ with a Gaussian-smoothed salt \& pepper-based noise image $SP_c$, simulating the random presence of larger particles such as colored dissolved matter:
	\begin{equation}
	I_c(\textbf{x})= uI^{sct}_c(\textbf{x}) + (1-u)SP_{c\{sp_{col}, pr_c,\sigma_c\}},
	\label{eq_3_4}
	\end{equation}
	where $u \in [0, 1]$ is an image weighting parameter, $sp_{col}$ is a base particle color, $pr_c$ is the probability of adding a particle on $SP$'s channel $c$, and $\sigma_c$ is the deviation of the Gaussian blur applied to $SP$'s channel $c$.
	
	\section{Proposed Method}
	\label{proposed_method}
	\subsection{Network Architecture}
\begin{figure}[!htb]
		\centering{\includegraphics[trim = 0.2cm 0.2cm 3.9cm 0.1cm, clip, scale = 0.22]{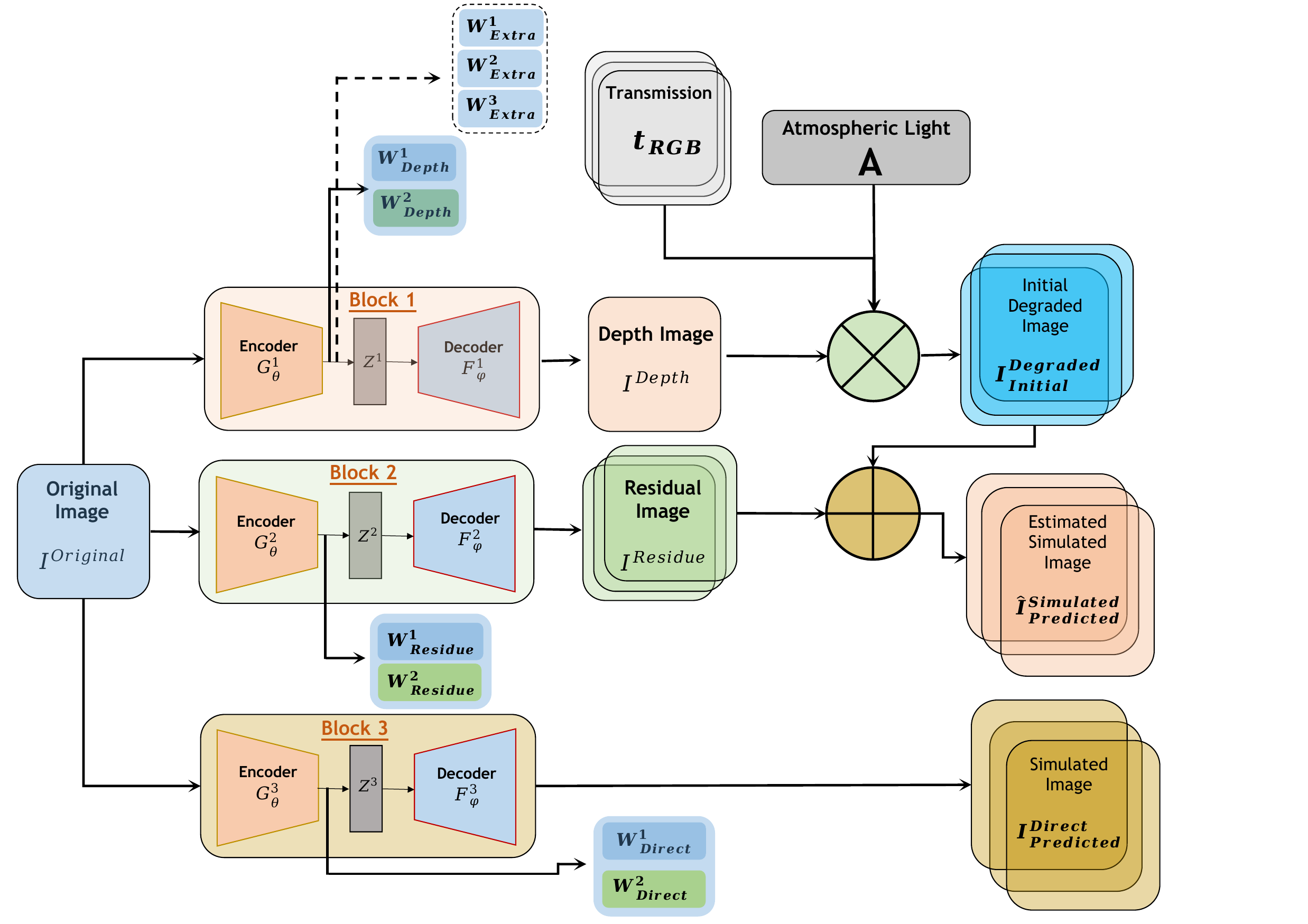}}
		\centering \caption{The proposed network architecture}
		\label{fig:archi_model}
	\end{figure}

	The complete network architecture is depicted in Fig.~\ref{fig:archi_model}. We have used three parallel \textit{encoder-decoder} models, corresponding to three branches of the global models: \emph{i)} simple degraded image formation model, using estimated depth image (see Equation~\eqref{eq_1}) \emph{ii)} model for residual learning and \emph{iii)} direct prediction of degraded image. For all these three networks, the input R,G,B image is encoded into feature vectors by using a DenseNet-169 \cite{Huang2017} network which is pretrained on ImageNet \cite{Deng2010}. Then this vector is then fed into a successive series of upsampling layers in order to construct the final depth map at half of the input resolution of the input image. These upsampling layers and their associated skip-connections form our \emph{decoder}. The proposed decoder model is simple and straightforward \cite{Ioffe2015}. Further architectural details are provided in the supplementary materials. 
	The complete network architecture is depicted in Fig.~\ref{fig:archi_model}. We have used three independent \textit{encoder-decoder} models in this network. The original clean image is taken as the input in all of these three \textit{encoder-decoder} models. The first \textit{encoder-decoder} is used to predict the gray level depth image $(I^{Depth})$. The second \textit{encoder-decoder} network is used to predict the $RGB$ \textit{residual} image $(I^{Residue})$ and the third \textit{encoder-decoder} network is used to directly predict the \textit{simulated underwater image} $(I^{Simulated}_{Predicted})$ from the original image. By using the estimated depth image $(I^{Depth})$, user given \emph{atmospheric light} $(A_c)$ and \emph{attenuation coefficients} ($\beta_c$, which is further used to calculate $t_c$; see Equation~\eqref{eq_2}), we compute the    
	initial model-based degraded image $(I_{Initial}^{Degraded})$ using Equations~\eqref{eq_1} and \ref{eq_2}. Now, by using $I_{Initial}^{Degraded}$ and $I^{Residue}$, we compute the \emph{estimated simulated image} $(\hat{I}_{Simulated}^{Predicted})$. We apply the third \textit{encoder-decoder} network also to directly estimate the \emph{simulated image} ($I^{Direct}_{Predicted}$) from the input of \emph{RGB} image ($I^{Original}$) only.	
	To sum up, this architecture is based on $3$ individual blocks of \emph{encoder-decoder} networks. The objective of the first network is to estimate the depth image which is further used to obtain an initial degraded image by applying the physics induced image formation model (see Equation~\eqref{eq_1}). The second branch learns the residual by capturing everything that is not modeled by Equation~\eqref{eq_1} in a data driven manner. The third branch directly predicts the ground truth image. 
	\subsection{Learning Loss Function}
	Any standard loss function for predicted image regression problems considers a discrepancy measure between the ground truth image and the predicted image. However, different considerations regarding the loss function can have very significant effects on the training speed and the overall performance of image estimation. Here, we have categorically analyzed the influence or significance of each of the three branches of our architecture on the overall performance of the network. 
	\subsubsection{Technique 1:} As a first approach, we use the initial two \emph{encoder-decoder} blocks (omitting the third or bottom most \emph{encoder-decoder} block) of the network (see Fig.~\ref{fig:archi_model}). Our objective here is firstly to define a loss function that balances between the predicted ``depth image'' $(I_{depth})$ by minimizing the difference of depth values while also penalizing distortions of high frequency details in the depth image domain (these details typically correspond to object boundaries in the scene). Secondly, we also minimize the difference between \emph{predicted simulated image} $(I^{Simulated}_{Predicted})$ and the manually calculated ground truth version of the same image.
	Thirdly, based on the predicted \emph{depth image}	$(I_{depth})$, user given \emph{attenuation coefficients} ($\beta_c$), which is further used to calculate the ``transmission matrix'' ($t_c$) and \emph{atmospheric light} $(A)$, we first generate the ``initial degraded image'' $(I_{Initial}^{Degraded})$, which is then combined with the estimated ``residual'' image i.e. $I^{Residue}$ to generate the ``estimated simulated image'' $\hat{I}_{Simulated}^{Predicted}$ (see Equation~\eqref{loss_eq_1}). 
	Hence, we need to take into account the correct estimation of $I_{depth}$, followed by the correct estimation of $I_{Initial}^{Degraded}$ which is the main contribution in the formation of $\hat{I}_{Simulated}^{Predicted}$. It is also equally important to properly estimate $I^{Residue}$ as it will represent the missing quantity between the targeted estimated image $\hat{I}_{Simulated}^{Predicted}$ and $I_{Initial}^{Degraded}$. That is why we add $I_{Initial}^{Degraded}$ and $I^{Residue}$ to obtain the $\hat{I}_{Simulated}^{Predicted}$. 
	
	To train using this configuration, we first define two separate loss functions i.e. $L_d$, $L_p$ and compute the final loss $L_{total} = L_d + L_p$. $L_d$ represents the loss corresponding to the depth image $(I_{depth})$ reconstruction, and $L_p$ represents the loss corresponds to the ``estimated simulated image'' $(\hat{I}_{Simulated}^{Predicted})$ reconstruction.
	$L_d$ can be defined as follows:
	\begin{equation}
	L_d(y_p, \hat{y}_p) = \lambda_1^y L_{depth}(y_p, \hat{y}_p) + \lambda_2^y L_{SSIM}(y_p, \hat{y}_p)
	\label{depth_equation}
	\end{equation}
	The depth term $L_{depth}$ is the point-wise $L1$ loss which is defined on the depth values :
	\noindent\begin{minipage}{.5\linewidth}
		\begin{equation}
		L_{depth}(y_p,\hat{y}_p) = \frac{1}{n} \sum_{p}^{n}|y_p - \hat{y}_p|
		\label{depth_1_eq}
		\end{equation}
	\end{minipage}%
	\begin{minipage}{.5\linewidth}
		\begin{equation}
		L_{SSIM}(y_p, \hat{y}_p) = \frac{1-SSIM(y, \hat{y})}{2}
		\label{ssim_1_eq}
		\end{equation}
	\end{minipage}
	
	The second term $L_{SSIM}$ uses the Structural Similarity (SSIM) \cite{Wang2004}, a common metric for image reconstruction tasks and it has been shown to be a good loss term for depth estimation by using CNNs \cite{Godard2017}. As SSIM $\leq$ $1$, we define a loss $L_{SSIM}$ as in Equation~\eqref{ssim_1_eq}. 
	Please note that, in Equation~\eqref{depth_equation}, we only have defined $\lambda_1^y$ and $\lambda_2^y$ as two weighting parameters which we have empirically set to $\lambda_1^y = \lambda_2^y = 0.1$. The inherited problem with such loss terms is that they tend to be larger when the ground-truth depth values are bigger. In order to resolve this issue, the reciprocal depth values are considered \cite{Huang2018a}, where the original depth map $y_{orig}$ is replaced by the target depth map $y=m/y_{orig}$; where $m$ is the maximum depth in the scene (e.g. $m=10$ meters for the NYU Depth v2 dataset). Our approach considers transforming the depth values and computing the loss in the log space \cite{Eigen2014a}, \cite{Ummenhofer2017}.   	
	The second loss $L_p$ is defined as:
	\begin{equation}
	L_p(q_p, \hat{q}_p) = \lambda_1^q ~\left[\frac{1}{n} \sum_{p}^{n}|q_p - \hat{q}_p|\right] + \lambda_2^q ~\left[ \frac{1-SSIM(m_p, \hat{q}_p)}{2} \right]
	\label{predicted_equation}
	\end{equation}
	where $q_p$ and $\hat{q}_p$ represents true and predicted ``estimated simulated image'' $(\hat{I}_{Simulated}^{Predicted})$.
	\begin{equation}
	\begin{array}{l}
	I^{Degraded}_{Initial}(\textbf{x}) = t_c(\textbf{x})I_{original}(\textbf{x})+(1-t_c(\textbf{x})).A_c \\ \\
	t_c(\textbf{x}) = \exp^{-\beta_c I_{Depth}(\textbf{x})}; 	\hspace{5mm}
	\hat{I}^{Simulated}_{Predicted}(\textbf{x}) = I^{Degraded}_{Initial}(\textbf{x}) + I^{Residue}
	\end{array}
	\label{loss_eq_1}
	\end{equation}
	Finally, the total loss is calculated as follows : 
	$L_{total} = L_d(y_p, \hat{y}_p) + L_p(q_p, \hat{q}_p)$. 
	\label{total_loss_equation}
	Furthermore, for this technique and the next ones, we also propose following two more variants to compute the loss function. 
	\paragraph{Variant 1 :} 
	\label{tech_1_variant_1}
	Instead of fixing the values of  $\lambda_1^q$ and $\lambda_2^q$ as $0.1$, we compute them automatically by using the same network, shown in Fig.~\ref{fig:archi_model}. After obtaining the pre-trained features from the last encoder block of \emph{DenseNet}, those are passed through two blocks of \emph{Fully Connected (FC) Layers}.
	These features are then flattened and reduced in dimension by passing through several linear layers with ReLU activations. Finally these features are passed through \emph{sigmoid} layer to obtain two weights (see Block-1 and Block-2 in Fig.~\ref{fig:archi_model}. Further architectural details are mentioned in supplementary material)\footnote{For this architectural configuration, we do not create the second branch from \textbf{Block 1}, corresponding to $W^1_{Extra}$, $W^2_{Extra}$, $W^3_{Extra}$ as we don't need these weights}. By using these automatic  weight values, the $L_d$ and $L_p$ is calculated as:
	\begin{equation}
	\begin{array}{l}
	L_d(y_p, \hat{y}_p) = w^1_{Depth} L_{depth}(y_p, \hat{y}_p) + (1- {w^1_{Depth}}) L_{SSIM}(y_p, \hat{y}_p) \\
	L_p(q_p, \hat{q}_p) = {w^1_{Residue}} ~\left[\frac{1}{n} \sum_{p}^{n}|q_p - \hat{q}_p|\right] + (1-{w^1_{Residue}}) ~\left[ \frac{1-SSIM(q_p, \hat{q}_p)}{2} \right]
	\\
	L_{total} = L_d(y_p, \hat{y}_p) + L_p(q_p, \hat{q}_p)
	\end{array}
	\label{tech_1_variant_1_equation}
	\end{equation}
	\paragraph{Variant 2 :} In addition with the weighted computation of $L_d(y_p, \hat{y}_p)$ and $L_p(q_p, \hat{q}_p)$ (according to Equation~\eqref{tech_1_variant_1_equation}), we also compute the total loss in the following manner. 
	\begin{equation}
	\begin{array}{l}
	L_{total} = {w^2_{Depth}} L_d(y_p, \hat{y}_p) + (1 - {w^2_{Depth}}) L_p(q_p, \hat{q}_p)
	\end{array}
	\label{tech_1_variant_2_equation}
	\end{equation}
	
	\subsubsection{Technique 2:} As a second approach, a loss term corresponding to the ``initial degraded'' image (computed by using Equation~\eqref{eq_1}) is added, compared to the total loss of Equation~\eqref{tech_1_variant_1_equation}. Along with the loss of depth image $(I_{Depth})$ and ``estimated simulated image'' $(\hat{I}^{Predicted}_{Simulated})$, we compute the loss (denoted as $L_t$) on $I^{Degraded}_{Inital}$ as: 
	\begin{equation}
	L_t(h_p, \hat{h}_p) = \lambda_1^h ~\left[\frac{1}{n} \sum_{p}^{n}|h_p - \hat{h}_p|\right] + \lambda_2^h ~\left[ \frac{1-SSIM(h_p, \hat{h}_p)}{2} \right]
	\label{haze_loss}
	\end{equation}
	where $h_p$ and $\hat{h}_p$ represents the true and predicted ``initial degraded'' image $({I}_{Initial}^{Degraded})$, and $\lambda_1^h$, $\lambda_2^h$ are set to $0.1$. Finally, the total loss is computed by: 
	\begin{equation}
	L_{total} = L_d(y_p, \hat{y}_p) + L_p(q_p, \hat{q}_p) + L_t(h_p, \hat{h}_p)
	\label{total_loss_equation_tech_2}
	\end{equation}
	As the \emph{variant 1} under this category, here also we compute the weighted (automatic) version of $ L_t(h_p, \hat{h}_p)$ in addition to the weighted version of $L_d(y_p, \hat{y}_p)$ and $L_p(q_p, \hat{q}_p)$ (as in Equation~\eqref{tech_1_variant_1_equation},) as:
	\begin{equation}
	\begin{array}{l}
	
	L_t(h_p, \hat{h}_p) = {w^2_{Depth}} ~\left[\frac{1}{n} \sum_{p}^{n}|h_p - \hat{h}_p|\right] + (1-{w^2_{Depth}}) ~\left[ \frac{1-SSIM(h_p, \hat{h}_p)}{2} \right]
	\\
	L_{total} = L_d(y_p, \hat{y}_p) + L_p(q_p, \hat{q}_p) + L_t(h_p, \hat{h}_p)
	\end{array}
	\label{total_loss_equation_tech_2_variant_1}
	\end{equation}
	As the \emph{variant 2} under this category, the total loss is computed as (see supplementary material for more details about the architecture).  
	\begin{dmath}
		L_{total} = {w^1_{Extra}} \times L_d(y_p, \hat{y}_p) + {w^2_{Extra}} \times L_p(q_p, \hat{q}_p)  +  [1-({w^1_{Extra}} + {w^2_{Extra}})] \times L_t(h_p, \hat{h}_p)
		\label{tech_2_variant_2_equation}
	\end{dmath}
	\subsubsection{Technique 3:} As the third modification, here we introduce an additional \emph{encoder-decoder} block to directly estimate the simulated image $(I^{Direct}_{Predicted})$ from the clean \emph{RGB} image. Hence, we compute a dedicated loss ($L_g$) for $(I^{Direct}_{Predicted})$ image only:
	\begin{equation}
	L_g(s_p, \hat{s}_p) = \lambda_1^s ~\left[\frac{1}{n} \sum_{p}^{n}|s_p - \hat{s}_p|\right] + \lambda_2^s ~\left[ \frac{1-SSIM(s_p, \hat{s}_p)}{2} \right]
	\label{direct_loss}
	\end{equation}
	where $s_p$ and $\hat{s}_p$ represents the true and predicted ``directly simulated'' image $({I}_{Initial}^{Degraded})$ by the network and the value of $\lambda_1^s$ and $\lambda_2^s$ are set to $0.1$. Finally, the total loss is: 
	\begin{equation}
	L_{total} = L_d(y_p, \hat{y}_p) + L_p(q_p, \hat{q}_p) + L_t(h_p, \hat{h}_p) + L_g(s_p, \hat{s}_p)
	\label{total_loss_equation_tech_3}
	\end{equation}
	As the \emph{variant 1} under this category, here also we compute the weighted (automatic) version of $L_g(s_p, \hat{s}_p)$ in addition to the weighted version of $L_d(y_p, \hat{y}_p)$, $L_p(q_p, \hat{q}_p)$ and $L_t(h_p, \hat{h}_p)$ (in the same manner as in Equation~\eqref{total_loss_equation_tech_2_variant_1}) in the following manner.
	\begin{equation}
	\begin{array}{l}
	
	L_g(s_p, \hat{s}_p) = {w^1_{Direct}} ~\left[\frac{1}{n} \sum_{p}^{n}|s_p - \hat{s}_p|\right] + (1-{w^1_{Direct}}) ~\left[ \frac{1-SSIM(s_p, \hat{s}_p)}{2} \right]
	\\
	L_{total} = L_d(y_p, \hat{y}_p) + L_p(q_p, \hat{q}_p) + L_t(h_p, \hat{h}_p) + L_g(s_p, \hat{s}_p)
	\end{array}
	\label{total_loss_equation_tech_3_variant_1}
	\end{equation}
	As the \emph{variant 2} under this category, the total loss is computed in the following manner (see supplementary material for more details about the architecture). 
	\begin{dmath}
		L_{total} = {w^1_{Extra}} \times L_d(y_p, \hat{y}_p) + {w^2_{Extra}} \times L_p(q_p, \hat{q}_p) + {w^3_{Extra}} \times L_t(h_p, \hat{h}_p) +  [1-({w^1_{Extra}} + {w^2_{Extra}} + {w^3_{Extra}} )] \times L_t(h_p, \hat{h}_p)
		\label{tech_3_variant_2_equation}
	\end{dmath}
	%
	\section{Experimental Results}
	\subsection{Datasets}
	We have used following two well known datasets which provide \emph{RGB-D} images. 
	\subsubsection{NYU Depth v2: } 
	This dataset provides images and depth maps for different indoor scenes, captured at the resolution of $640 \times 480$ \cite{Silberman2012}. 
	For this work, we have used a subset of $50k$ images which we obtained from \cite{Alhashim2018}. 
	The depth map has an upper bound of $10$ meters. Like in \cite{Alhashim2018}, our method also produces the predictions at the half of the input resolution i.e. at $320 \times 240$ and here also we do not crop any of the input image-depth map pairs even though they contain missing pixels, due to the preprocessing for distortion correction. We have used $96\%$ i.e. $48650$ images for training and the remaining $2030$ images are used for testing purposes.  
	\subsubsection{Make3D: } 
	This dataset contains $534$ RGB-depth pairs, split into $400$ pairs for training and $134$ for testing. The RGB images are provided at high resolution while the available depth maps are comparatively at very low resolution. Therefore, the data is resized into $460 \times 345$ as proposed in \cite{Saxena2005} \cite{Saxena2007}. We used the same data reading and processing protocol as in \cite{Gur2019a} and the results are evaluated by using depth cap of $0-80$.
	\subsection{Evaluation}
	We use standard metrics \cite{Eigen2014a} to quantitatively compare our method against state-of-the-art techniques. These error metrics are defined as:
	\begin{enumerate}[i.]
		\item Average relative error (rel): $\frac{1}{n} \sum_{p}^{n} \frac{|y_p - \hat{y}_p|}{y}$
		\item Root mean square error (rms): $\sqrt{\frac{1}{n} \sum_{p}^{n} (y_p - \hat{y}_p)^2}$
		\item Average ($\log_{10}$) error: $\frac{1}{n} \sum_{p}^{n}|\log_{10}{(y_p)} - \log_{10}{(\hat{y}_p)}|$
		\item Threshold accuracy ($\delta_i$): \% of $y_p$ s.t. $max(\frac{y_p}{\hat{y}_p}, \frac{\hat{y}_p}{y_p}) = \delta < Thersh$ for $Thresh = 1.25, 1.25^2, 1.25^3;$
	\end{enumerate}
	Where $y_p$ is a pixel in depth image $y$, $\hat{y}_p$ is a pixel in the predicted depth image $\hat{y}$ and $n$ is the total number of pixels for each depth image.
	
	In Tables~\ref{MTL_accuracy_2}, we report the accuracy of the NYU-V2 dataset. The performance of $9$ variants of the proposed methods are compared with $6$ other relevant techniques. The \emph{variant-1} for each technique performs slightly better than the core technique (i.e. \emph{Technique-1 Variant-1} has performed better than \emph{Technique-1} itself). Whereas \emph{variant-2} outperforms the \emph{variant-1} by a large margin. 
	These results signifies that performing the weighted combination of the contribution from different entities in the total loss (see Equation~\eqref{tech_1_variant_2_equation}) strongly improves the results compared to considering equal and full contribution of each term. 
	\vspace{-0.5cm}
	\begin{table}[]
		\small{
			\caption{Comparisons of different methods on the NYU Depth v2 dataset }
			\centering
			\begin{tabular}{|P{4.3cm}|P{1.2cm} P{1.2cm} P{1.2cm} || P{1.2cm} P{1.2cm} P{1.2cm}|}
				\hline
				
				\textbf{Method} & $\pmb {\delta_1} \uparrow$ & $\pmb {\delta_2} \uparrow $ & $\pmb {\delta_3} \uparrow$ & $\pmb {rel} \downarrow $ & $\pmb {rms} \downarrow $ & $ \pmb{\log_{10}} \downarrow$  \\ \hline
				
				\mycc \textit{Proposed:} Technique-1 & \mycc 0.505 &  \mycc 0.747 & \mycc 0.890 & \mycc 0.614 & \mycc 0.176 & \mycc 0.150 \\
				\hline
				\mycc \textit{Proposed:} Technique-1 Variant-1 & \mycc 0.512 &  \mycc 0.750 & \mycc 0.892 & \mycc 0.633 & \mycc 0.174 & \mycc 0.150 \\
				\hline
				\mycc \textit{Proposed:} Technique-1 Variant-2 & \mycc \bf{0.769} &  \mycc \bf{0.947} & \mycc \bf{0.986} & \mycc \bf{0.165} & \mycc \bf{0.109} & \mycc \bf{0.065} \\
				\hline
				\mybb \textit{Proposed:} Technique-2 & \mybb 0.507 &  \mybb 0.748 & \mybb 0.888 & \mybb 0.610 & \mybb 0.172 & \mybb 0.150 \\	
				\hline
				\mybb \textit{Proposed:} Technique-2 Variant-1 & \mybb 0.509 &  \mybb 0.736 & \mybb 0.885 & \mybb 0.622 & \mybb 0.173 & \mybb 0.150 \\	
				\hline
				\mybb \textit{Proposed:} Technique-2 Variant-2 & \mybb \bf{0.779} &  \mybb \bf{0.945} & \mybb \bf{0.984} & \mybb \bf{0.163} & \mybb \bf{0.105} & \mybb \bf{0.065} \\	
				\hline
				\myrr \textit{Proposed:} Technique-3 & \myrr 0.510 &  \myrr 0.743 & \myrr 0.883 & \myrr 0.601 & \myrr 0.173 & \myrr 0.150 \\	
				\hline
				\myrr \textit{Proposed:} Technique-3 Variant-1 & \myrr 0.514 &  \myrr 0.748 & \myrr 0.895 & \myrr 0.627 & \myrr 0.173 & \myrr 0.149 \\	
				\hline
				\myrr \textit{Proposed:} Technique-3 Variant-2 & \myrr \textit{\textbf{0.792}} &  \myrr \textit{\textbf{0.952}} & \myrr \textit{\textbf{0.987}} & \myrr \textit{\textbf{0.152}} & \myrr \textit{\textbf{0.100}} & \myrr \textit{\textbf{0.060}} \\
				\hline
				Encoder-Decoder Model \cite{Alhashim2018} &  0.269 &  0.556 & 0.776 & 1.121 & 0.2345 & 0.231 \\	
				\hline
				Pix2Pix \cite{Isola2016} &  0.743 &  0.900 & 0.957 & 0.204 & 0.069 & 0.080 \\	
				\hline
				CycleGAN \cite{Zhu2017a} & 0.227 &  0.418 & 0.572 & 1.242 & 0.303 & 0.315 \\
				\hline
				UNIT \cite{Liu2017} & 0.220 &  0.402 & 0.559 & 1.194 & 0.307 & 0.334 \\
				\hline
				MUNIT \cite{Huang2018b} & 0.233 &  0.364 & 0.482 & 1.394 & 0.341 & 0.349 \\
				\hline
				DRIT \cite{Lee2018a} & 0.246 &  0.451 & 0.606 & 1.223 & 0.311 & 0.298 \\
				\hline
				\hline
			\end{tabular}
			\label{MTL_accuracy_2}
		}
	\end{table}	
	\begin{table}[]
		\small{
			\caption{Comparisons of different methods on the Make-3D dataset }
			\centering
			\begin{tabular}{|P{4.3cm}|P{1.2cm} P{1.2cm} P{1.2cm} || P{1.2cm} P{1.2cm} P{1.2cm}|}
				\hline
				
				\textbf{Method} & $\pmb {\delta_1} \uparrow$ & $\pmb {\delta_2} \uparrow $ & $\pmb {\delta_3} \uparrow$ & $\pmb {rel} \downarrow $ & $\pmb {rms} \downarrow $ & $ \pmb{\log_{10}} \downarrow$  \\ \hline
				
				\mycc \textit{Proposed:} Technique-1 & \mycc 0.359 &  \mycc 0.603 & \mycc 0.757 & \mycc 1.613 & \mycc 0.284 & \mycc 0.226 \\
				\hline
				\mycc \textit{Proposed:} Technique-1 Variant-1 & \mycc 0.346 &  \mycc 0.608 & \mycc 0.767 & \mycc 1.55 & \mycc 0.267 & \mycc 0.223 \\
				\hline
				\mycc \textit{Proposed:} Technique-1 Variant-2 & \mycc 0.355 &  \mycc 0.617 & \mycc 0.772 & \mycc \textbf{1.527} & \mycc \textbf{0.262} & \mycc \textbf{0.220} \\
				\hline
				\mybb \textit{Proposed:} Technique-2 & \mybb \textbf{0.383} &  \mybb \textbf{0.627} & \mybb 0.771 & \mybb 1.760 & \mybb 0.292 & \mybb 0.223 \\	
				\hline
				\mybb \textit{Proposed:} Technique-2 Variant-1 & \mybb 0.366 &  \mybb 0.615 & \mybb 0.769 & \mybb 1.59 & \mybb 0.270 & \mybb 0.221 \\	
				\hline
				\mybb \textit{Proposed:} Technique-2 Variant-2 & \mybb 0.362 &  \mybb 0.625 & \mybb \textbf{0.773} & \mybb 1.60 & \mybb 0.269 & \mybb \textbf{0.220} \\	
				\hline
				\myrr \textit{Proposed:} Technique-3 & \myrr \textbf{0.379} &  \myrr 0.621  & \myrr 0.771  & \myrr 1.648 & \myrr 0.281  & \myrr \textbf{0.220} \\	
				\hline
				\myrr \textit{Proposed:} Technique-3 Variant-1 & \myrr 0.330 &  \myrr 0.604 & \myrr 0.766 & \myrr 1.905 & \myrr 0.271 & \myrr 0.232 \\	
				\hline
				\myrr \textit{Proposed:} Technique-3 Variant-2 & \myrr 0.365 &  \myrr \textbf{0.626}  & \myrr \textbf{0.778}  & \myrr 1.589 & \myrr \textbf{0.263} & \myrr \textbf{0.218} \\
				\hline
				Encoder-Decoder Model \cite{Alhashim2018} &  0.331 &  0.619  & \textit{\textbf{0.808}}  & \textbf{1.485} & \textit{\textbf{0.242}} & \textit{\textbf{0.211}} \\
				\hline
				Pix2Pix \cite{Isola2016} &  0.239 &  0.458  & 0.638  & \textit{\textbf{1.367}} & 0.301 & 0.264 \\
				\hline
				CycleGAN &  \textit{\textbf{0.654}} &  \textit{\textbf{0.698}}  & 0.741  & 1.865 & 0.760 & 0.567 \\
				\hline
				\hline
			\end{tabular}
			\label{MTL_accuracy_3}
		}
		\vspace{-0.5cm}
	\end{table}	
	Now, if we compare \emph{Technique-1}, \emph{Technique-2} and their variants, there is only minute difference in results within these approaches. Compared to \emph{Technique-1}, in \emph{Technique-2} the additional loss (i.e. $L_t(h_p, \hat{h}_p)$) of ``initial degraded'' (i.e. $I_{Initial}^{Degraded}$) image is added. As we have already used the depth image loss ($L_{d}$) in \emph{Technique-1} and \emph{Technique-2}, adding the $L_t$ loss for $I_{Initial}^{Degraded}$ image, which is computed by using the depth image (i.e. $I_{Depth}$) and other user given constants i.e. $I_{Original}$, $\beta_c$ and $A_c$ (see Equation~\eqref{loss_eq_1}) does not make much difference in terms of loss level contribution. Furthermore, in \emph{Technique-3} we add the additional loss $L_g$, corresponding to $I_{Predicted}^{Direct}$ image which also inherently adds computational burden, related to the third block of \emph{encoder-decoder} network (see Fig.~\ref{fig:archi_model}). But adding this extra loss of $L_g$ could only slightly improve the performance. Hence, we should find a good trade-off between leveraging slight improvement in accuracy compared to the additional computational burden. 
	\begin{figure*}[bht!]
		\centering	
		\begin{subfigure}[b]{\textwidth}
			\centering {\includegraphics[trim = 0.0cm 8.6cm 0.0cm 0.0cm, clip, scale = 0.18]{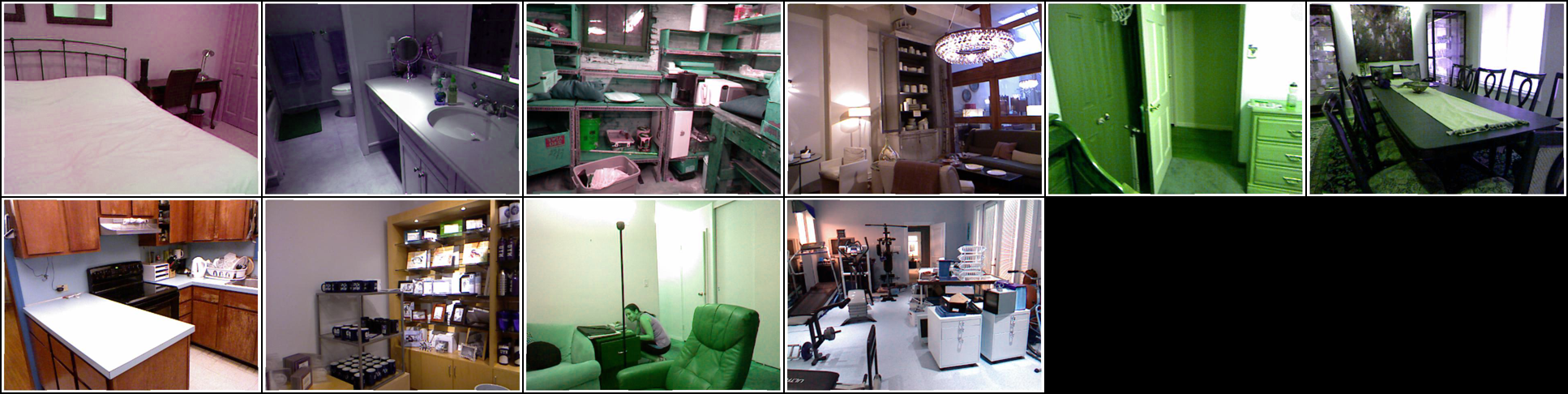} }
			\centering
			\vspace{-5mm}
			\caption{}
			\label{fig:technique_1_orig_half}
		\end{subfigure}			
		\begin{subfigure}[b]{\textwidth}
			\centering {\includegraphics[trim = 0.0cm 8.6cm 0.0cm 0.0cm, clip, scale = 0.18]{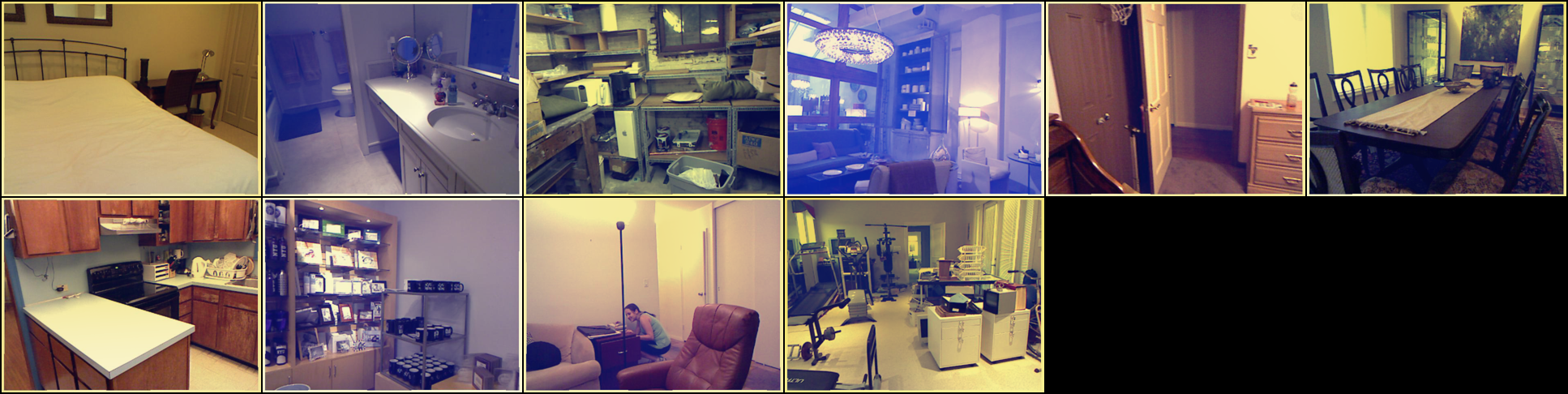} }
			\centering
			\vspace{-5mm}
			\caption{}
			\label{fig:technique_1_haze_GT_half}
		\end{subfigure} 
		\begin{subfigure}[b]{\textwidth}
			\centering {\includegraphics[trim = 0.0cm 8.6cm 0.0cm 0.0cm, clip, scale = 0.18]{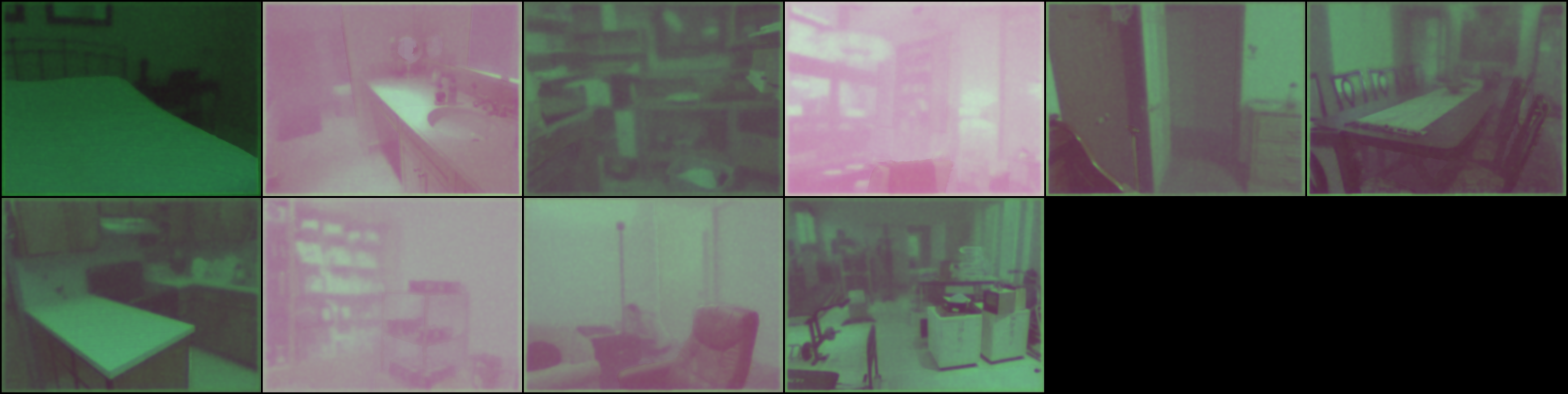} }
			\centering
			\vspace{-5mm}
			\caption{}
			\label{fig:technique_1_complex_GT_half}
		\end{subfigure}
		\begin{subfigure}[b]{\textwidth}
			\centering {\includegraphics[trim = 0.0cm 8.6cm 0.0cm 0.0cm, clip, scale = 0.18]{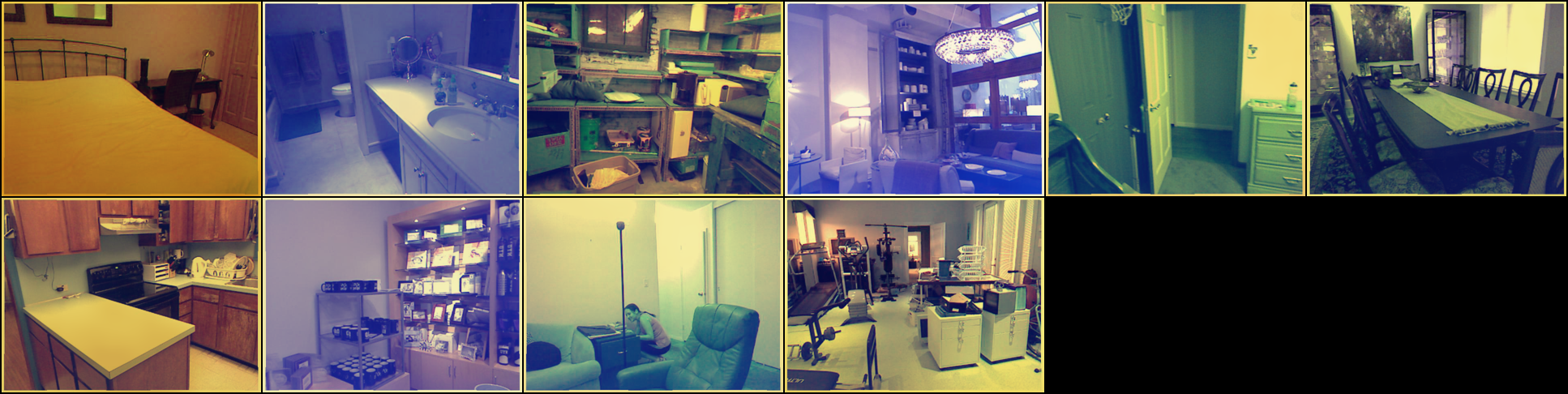} }
			\centering
			\vspace{-5mm}
			\caption{}
			\label{fig:technique_1_haze_Pred_half}
		\end{subfigure}
		\begin{subfigure}[b]{\textwidth}
			\centering {\includegraphics[trim = 0.0cm 8.6cm 0.0cm 0.0cm, clip, scale = 0.18]{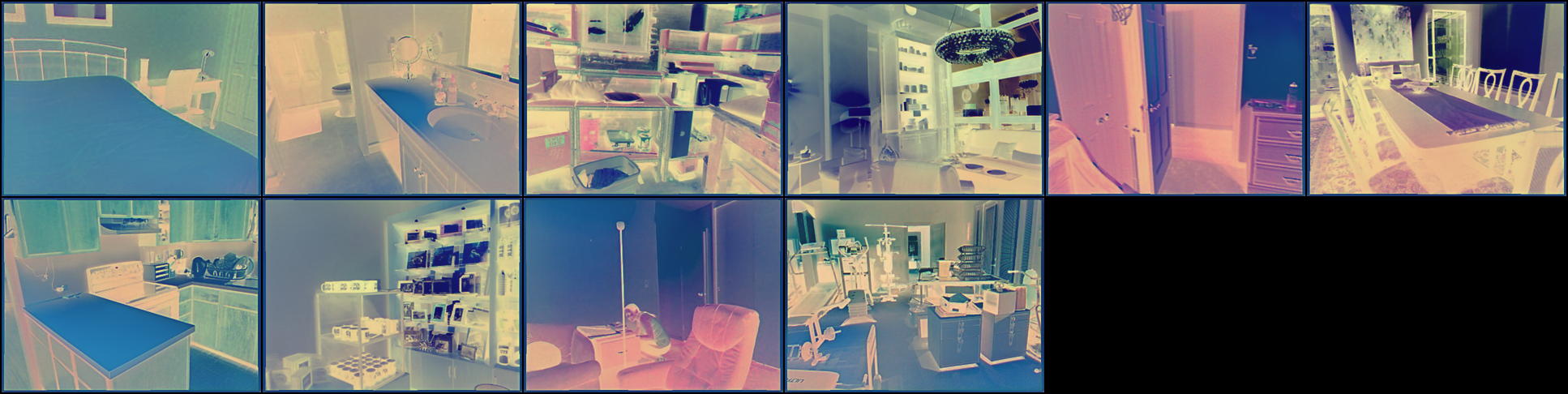} }
			\centering
			\vspace{-5mm}
			\caption{}
			\label{fig:technique_1_Residue_Pred_half}
		\end{subfigure}
		\begin{subfigure}[b]{\textwidth}
			\centering {\includegraphics[trim = 0.0cm 8.6cm 0.0cm 0.0cm, clip, scale = 0.18]{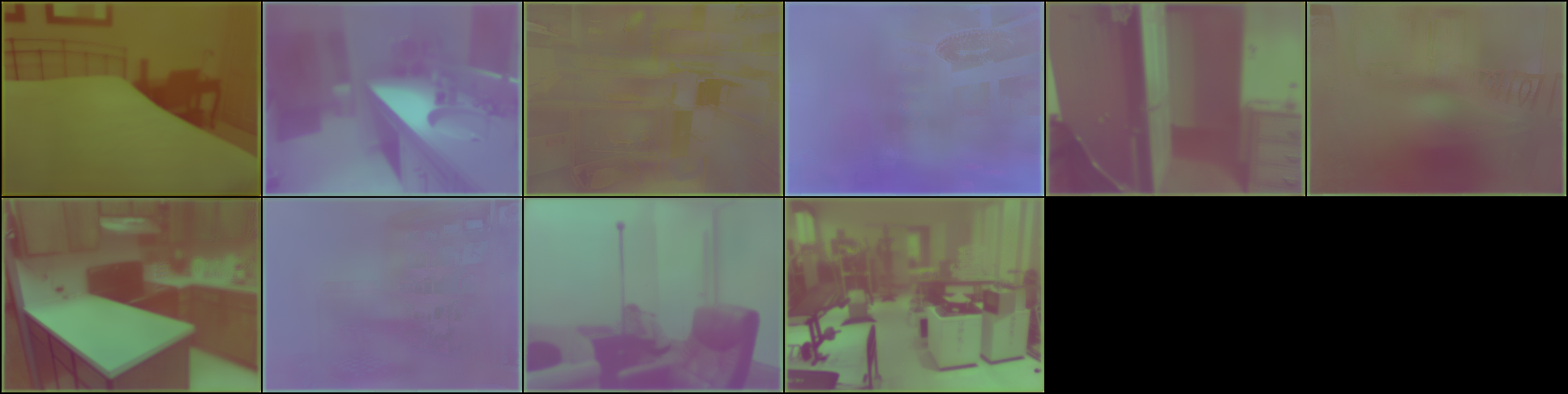} }
			\centering
			\vspace{-5mm}
			\caption{}
			\label{fig:technique_1_Complex_Pred_half}
		\end{subfigure}
		\caption[endProcess]{ \textbf{Qualitative Measures obtained by \textit{Proposed} Technique 1 :} (a) Original \emph{RGB} images (i.e. $J_c(\mathbf{x})$ in Equation~\eqref{eq_1}) (b) Ground Truth of ``Initial Degraded Image'' (i.e. $I_c(\mathbf{x})$ in Equation~\eqref{eq_1}) (c)  Ground Truth of ``Simulated Underwater Image'' (i.e. $I^{sct}_c(\textbf{x})$ in Equation~\eqref{eq_3_3}) (d) Predicted ``Initial Degraded Image'' (i.e. $I^{Degraded}_{Initial}$ in Fig.~\ref{fig:archi_model}) (e) Predicted ``Residual Image'' (i.e. $I^{Residue}$ in Fig.~\ref{fig:archi_model}) (f) Predicted ``Simulated Underwater Image'' (i.e. $\hat{I}^{Simulated}_{Predicted}$ in Fig.~\ref{fig:archi_model}).  }
		\label{fig:all_single_task}
		\vspace{-0.5cm}
	\end{figure*}
	The values of $\delta_1$, $\delta_2$ and $\delta_3$ (which count the number of pixels which are similar to each other between the ground truth and predicted image with respect to three different threshold values) shows that the \emph{variant-2} of all the three techniques have shown strong improvement in accuracy compared the core technique and \emph{variant-1}. However, the error metrics i.e. \emph{rel}, \emph{rms} and $\log_{10}$ (which calculates the pixel level errors between the ground truth and predicted image) shows that \emph{variant-2} of all the three techniques have shown strong improvement in accuracies. 
	
	The proposed techniques are compared with several other relevant image-to-image translation approaches. As the first such technique, we use only a single \emph{encoder-decoder} model (see third block in Fig.\ref{fig:archi_model}) similar to the technique in  \cite{Alhashim2018} to directly generate the degraded image from a \emph{RGB} image. Except Pix2Pix \cite{Isola2016}, all other comparable techniques do not perform well enough. Although Pix2Pix \cite{Isola2016} could perform better than other state-of-the-art techniques, this image-to-image translation approach is simply based on classical \emph{conditional-adversarial loss} and \emph{L1} loss which does not incorporate any physical image degradation model/equation like our proposed technique. The top $3$ results for each metric is noted in bold whereas the best result is mentioned in bold-italic style in Table.\ref{MTL_accuracy_2} and Table.\ref{MTL_accuracy_3}.   
	
	A few examples of the proposed \emph{Technique 1} is shown in Fig.~\ref{fig:all_single_task} where we can observe that to a good extent the proposed technique is able to generate ``initial degraded image'' ($I^{Degraded}_{Initial}$) (see Fig.~\ref{fig:technique_1_haze_Pred_half}) with respect to the ground truth, as shown in Fig.~\ref{fig:technique_1_haze_GT_half}. There are substantial differences between the ``simulated underwater image'' ($I_c^{sct}$) and $I^{Degraded}_{Initial}$ image (see Fig.~\ref{fig:technique_1_complex_GT_half} and Fig.~\ref{fig:technique_1_haze_GT_half}). The ``haze image'' explains some of the physical degradation in an interpretable way, but needs to be complemented by the data-driven residual. The proposed model can successfully capture this substantial difference i.e. the residual image is shown in Fig.~\ref{fig:technique_1_Residue_Pred_half}. By using these $I^{Degraded}_{Initial}$ and $I^{Residue}$, we can successfully predict the ``Simulated Underwater Image'' ($\hat{I}^{Simulated}_{Predicted}$), shown in Fig.~\ref{fig:technique_1_Complex_Pred_half}. Still, we are not always able to correctly reconstruct the colors perfectly but we can simulate the strong underwater blurry degradation effects.  
		
	It is difficult to draw any substantial conclusion from the results of Make-3D dataset (as is very small and the depth image resolution is quite low, compared to \emph{RGB} image, making it necessary to interpolate the depth image which highly degrades its quality), shown in Table~\ref{MTL_accuracy_3}. The same phenomenon is also visible that \emph{variant-2} has performed either better or very similar to it's counterparts i.e. the core technique and \emph{variant-1} for all of the three proposed techniques. By observing the weaker performance of other state-of-the-art methods from Table.~\ref{MTL_accuracy_2}, here we only have tested the simple \emph{encoder-decoder} model and most relevant as well as well known \emph{Pix2Pix}, \emph{CycleGAN} networks. The performance of these techniques are either close (e.g. \emph{encoder-decoder} network) or fall behind (e.g. \emph{Pix2Pix} model) the proposed techniques. Moreover, for certain metric (i.e. $\delta_1$ and $\delta_2$), \emph{CycleGAN} has outperformed others. Most importantly, as mentioned before that none of these technique incorporate any physical image degradation model/equation and are as interpretable as the proposed methods.  
	\section{Discussion}
	An interesting byproduct of our approach is that our trained model enables us to solve the inverse problem of underwater image restoration. Indeed, suppose we want to obtain a clean image $J_c$ from a degraded underwater image $y$. A natural way of carrying out this task is to minimize a mean squared error between the output of a known forward physical model $f$ and the real underwater image:
	\begin{equation*}
	\underset{J_c,\theta}{\textrm{arg min}}\ ||f(J_c,\theta) - y||^{2}
	\end{equation*}
	w.r.t. the input parameters of the model $\theta$, in our case veiling light, attenuation coefficients, depth image, and also the clean image $J_c$ itself; depth image could be estimated from degraded underwater image $y$ using a learned model. The resulting optimization problem requires computing the derivatives of $f$: $\frac{\partial f}{\partial \theta}$ and $\frac{\partial f}{\partial J_c}$ . Analytic derivatives are typically cumbersome to obtain for intricate physical models; hence leveraging automatic differentiation tools (e.g. Pytorch, Tensorflow etc.) is necessary. However, this requires $f$ to be perfectly known, and implemented in a modern automatic differentiation package,  \cite{nonnenmacher2021deep}. Here, the complex image formation model needs to be both known and differentiable. In real scenarios, a generative physical model is often unknown or badly known. Even when it is known, it is often implemented in languages that do not support automatic differentiation (e.g. C++, Fortran). With our approach, we obtain a deep learning-based emulator \cite{nonnenmacher2021deep}, in Pytorch: obtaining the derivatives of this model w.r.t to any of its inputs is straightforward thanks to it's automatic differentiation property. Moreover, the emulator does not require the knowledge of the underlying governing equations, its only task is to reproduce the desired outputs, even in a black-box fashion. Therefore, implementing algorithms to solve inverse problems which require to optimize over the model's input or parameters is easy with our trained model, and will be a basis for our future work.
	
	\section{Conclusion}
	In this paper, we have proposed a physics-informed and data-driven deep learning architecture to simulate the effect of underwater image degradation. We proposed a complex image formation model to create a simulated dataset from any RGB-depth available dataset. We proposed to inform our network with a simple haze image formation model that is able to account for simple image degradations, provided a good estimate of the depth image can be obtained. This image, as well as a residual image that captures the missing physics directly from data are obtained via DenseNet encoding-decoding blocks. Different losses are designed in order to estimate each component as well as their weighting parameters; which are obtained automatically. We obtain an emulator of this physical phenomenon that paves the way to obtaining differentiable and efficient emulators of complex physical models in other scenarios. We have shown on two datasets that our approach outperforms classical physics-ignorant deep learning models suited for image to image translation tasks. Future work will revolve around exploiting the interpretability and differentiability of the model to solve the inverse problem of image restoration. 
	
	\clearpage
	\bibliographystyle{splncs04}
	\bibliography{library}

	\title{Supplementary Materials}
	\maketitle
	
	\section{More Details on Image Formation Model}
	\subsection{Classical Simulation Model}
	The classical image formation model (developed in \cite{Schechner2005}) is written in Equation~\eqref{eq_1_ver_1}. The image intensity ($\textbf{x}$) at each pixel is composed of two components: the attenuated signal and the veiling light. 
	\begin{equation}
		I_c(\textbf{x}) = t_c(\textbf{x})J_c(\textbf{x})+(1-t_c(\textbf{x})).A_c
		\label{eq_1_ver_1}
	\end{equation}
	where bold letters denotes vectors, $\textbf{x}$ is the pixel coordinate, $I_c$ is the acquired image value in the color channel $c$, $t_c$ is the transmission of the color channel, and $J_c$ is the object radiance or clean image. The global veiling-light/atmospheric light component $A_c$ is the scene value in the areas with no objects $(t_c=0, \forall c \in \{R, G, B\})$. The transmission depends on the object's distance $z(\textbf{x})$ and the attenuation coefficient of the medium for each channel i.e. $\beta_c$:
	\begin{equation}
		t_c(\textbf{x}) = \exp^{-\beta_c z(\textbf{x})}
		\label{eq_2_ver_1}
	\end{equation}
	In the ocean, the attenuation of red colors can be an order of magnitude larger than the attenuation of blue and green \cite{Mobley2016}. Hence, contrary to the common assumption in single image dehazing, the transmission $\textbf{t} = (t_R, t_G, t_B)$ is wavelength dependent.
	
	\subsection{Water Attenuation}
	The attenuation of light in underwater is not constant and varies with the change in geography, seasons and climate. The attenuation coefficient ($\beta$) is dependent on wavelength of various water types. For clear open waters, the longest wavelength of visible light is first absorbed, resulting in deep blue colors to the eye. Waters near to the shore contain more suspended particles than the central ocean waters which scatter light and make coastal waters less clear than open waters. Moreover, the absorption of shortest wavelengths is stronger, thus the green wavelength reaches deeper than the other wavelengths. 
	Based on the water clarity, Jerlov \cite{Jerlov1976a} proposed a classification scheme for oceanic waters where open ocean waters are classified into class \textbf{I}, \textbf{IA}, \textbf{IB}, \textbf{II} and \textbf{III}. He also defined the water type $\textbf{1}$ through $\textbf{9}$ for coastal waters. Type \textbf{I} is the clearest and type \textbf{III} is the most turbid open ocean water. Similarly, for coastal water, type $\textbf{1}$ is the clearest and type $\textbf{9}$ is the most turbid. We use three attenuation coefficients: i.e. $\beta_R, \beta_G, \beta_B$ corresponding to $R$$G$$B$ channels for our work.
	
	\begin{figure}[!h]	       
		\begin{subfigure}[b]{0.49\linewidth}
			\centering{\includegraphics[trim = 0.2cm 0.2cm 0.1cm 0.4cm, clip,  scale = 0.18]{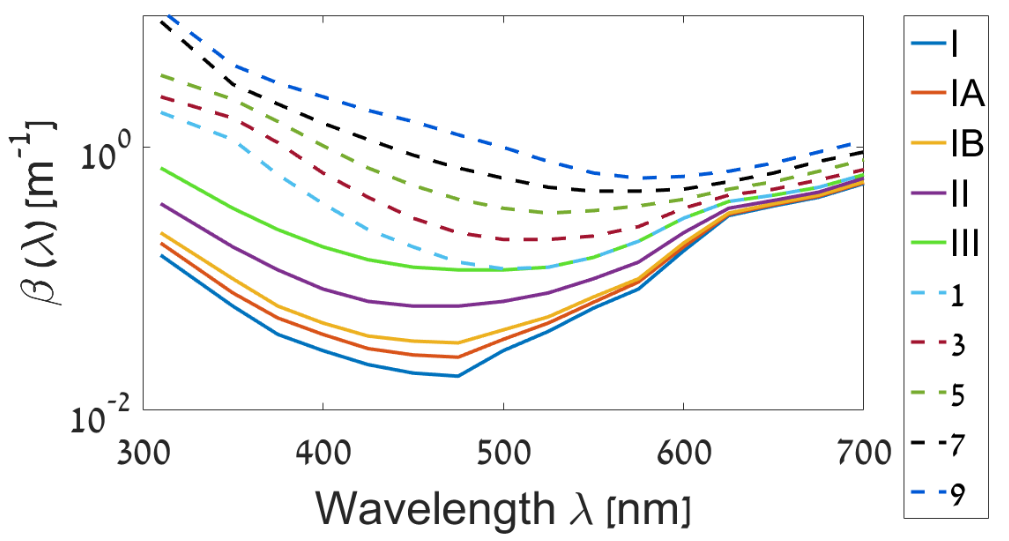}}
			\caption{}
			\label{fig:archi_resnet_single}
		\end{subfigure}
		\begin{subfigure}[b]{0.49\linewidth}
			\centering{\includegraphics[trim = 0.2cm 0.3cm 0.6cm 0.4cm, clip,  scale = 0.19]{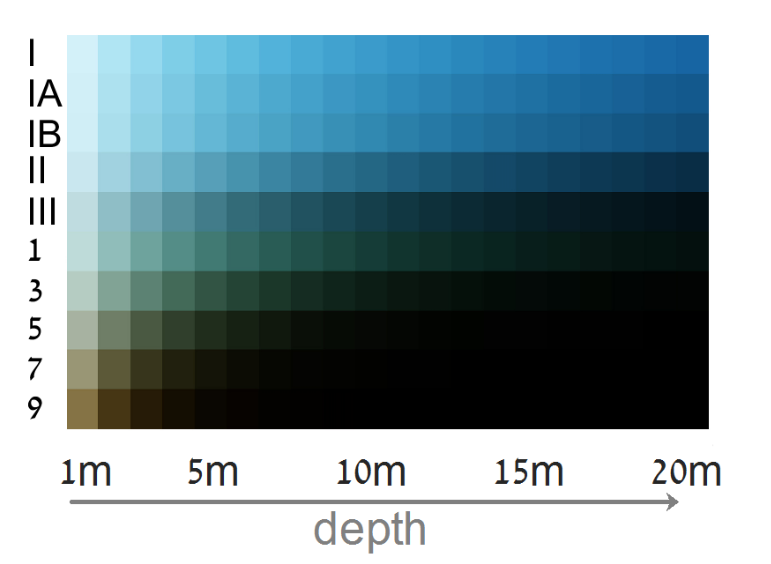}}
			\caption{}
			\label{fig:archi_resnet_multiple}
		\end{subfigure}	
		\caption{\small{(a) The attenuation coefficients ($\beta$) of Jerlov water types. The solid lines corresponds to open ocean water types while the dashed lines mark coastal water types. (b) For the case of different water types, the simulation of the appearance of white surface, viewed at depth of $1-20$m. (figures are taken from \cite{Berman2017})} 
		}
		\label{fig:archi_resnet}
	\end{figure}
	
	The Fig.~\ref{fig:archi_resnet_single} shows the attenuation coefficient ($\beta$) dependency on wavelength of various water types. Whereas, Fig.~\ref{fig:archi_resnet_multiple} shows an RGB simulation of the appearance of a perfect white surface viewed at different depths in different water types. A common notion is followed that red colors gets attenuated faster than blue/green colors only in the case of ocean water types. Based on three color channel i.e. R,G,B, we are interested in three corresponding attenuation coefficients: i.e. $\beta_R, \beta_G, \beta_B$ in order to correct/de-noise the image. In our work, we use only Jerlov water types to constrain the space of attenuation coefficients in the RGB domain.

	\section{Network Architecture}
	Our encoder-decoder network is based on \emph{DenseNet-169} \cite{Huang2017} where top most layer related to original ImageNet classification task is removed. Here, we use a pretrained \emph{DenseNet-169} model, which was trained on \emph{ImageNet} dataset \cite{imagenet_cvpr09}. The encoder structure is shown in Table.~\ref{MTL_accuracy_4} where the initial image of size $b \times 3 \times 480 \times 640$ (where $b$ is the batch size), is sequentially passed through several layers, which are mentioned in each row of the Table.~\ref{MTL_accuracy_4}. The size of the image gets gradually decreased by passing through each layer of the network and output size from each layers are mentioned in $2^{nd}$ column of the Table.~\ref{MTL_accuracy_4}. For more details about the \emph{DenseNet-169} network, please see \cite{Huang2017}. Please note that we have used pretrained model to easily obtain the image features from every \emph{DenseNet-169} network layers. 
	
	\begin{table}[]
		\small{
			\caption{DenseNet-169 based encoder model }
			\centering
			\begin{tabular}{|P{1cm}||P{2.9cm}|P{2.8cm}| P{4.8cm}|}
				\hline
				
				\textbf{Sl. No.} & \textbf{Layers} & \textbf{Output Size} &  \textbf{DenseNet-169}   \\ \hline
				$1$ & Input Image &	$ b \times 3 \times 480 \times 640 $   & $\times $ \\
				\hline			 
				$2$ & Convolution &	$ b \times 64 \times 240 \times 320 $   & No. of filters : $64$; Conv: $7 \times 7$; Stride: $2 \times 2$; Padding: $3 \times 3$ \\
				\hline
				$3$ & Batch Norm &	$ b \times 64 \times 240 \times 320 $   & $\epsilon = 10 ^{-5}$; momentum = $0.1$  \\
				\hline
				$4$ & ReLU Activation &	$ b \times 64 \times 240 \times 320 $   &  \\
				\hline
				$5$ & Max Pooling & $ b \times 64 \times 120 \times 160 $   &  Kernel Size: $3 \times 3$; Stride: $2 \times 2$; Padding: $1 \times 1$;  Dilation: $1 \times 1$   \\
				\hline
				\vspace{0.5cm}
				$6$ & \vspace{0.5cm} Dense Block - $1$ & \vspace{0.5cm} $ b \times 256 \times 120 \times 160 $    & \[ \left[  \frac{  ~~\text{Conv:~}1 \times 1 } { ~~\text{Conv:~}3 \times 3  } \right] \times 6\]
				\\
				\hline
				\multirow{2}*{$7$} &  \multirow{2}*{Transition Layer - $1$}  & $ b \times 128 \times 120 \times 160 $  & \text{Conv:}    $1 \times 1$ \\\cline{3-4}
				&   &	 $ b \times 128 \times 60 \times 80 $  &   \text{Average Pool : } $2 \times 2 $; Stride: $2 \times 2$  \\
				\hline
				\vspace{0.5cm}
				$8$ & \vspace{0.5cm} Dense Block - $2$ & \vspace{0.5cm} $ b \times 512 \times 60 \times 80 $ &    \[ \left[  \frac{  \text{Conv:~~} 1 \times 1 } {\text{Conv:}~~ 3 \times 3  } \right] \times 12\]
				\\
				\hline
				\multirow{2}*{$9$} &  \multirow{2}*{Transition Layer - $2$}  & $ b \times 256 \times 30 \times 40 $  & \text{Conv:}~~$1 \times 1$ \\\cline{3-4}
				&   &	 $ b \times 256 \times 30 \times 40 $  &   \text{Average Pool : } $2 \times 2 $; Stride: $2 \times 2$  \\
				\hline
				\vspace{0.5cm}
				$10$ & \vspace{0.5cm} Dense Block - $3$ &  \vspace{0.5cm} $ b \times 1280 \times 30 \times 40 $	  &    \[ \left[ \frac{ \text{Conv:~~} 1 \times 1 } {\text{Conv:~~} 3 \times 3  } \right] \times 32\]
				\\
				\hline
				
				\multirow{2}*{$11$} &  \multirow{2}*{Transition Layer - $3$}  & $ b \times 640 \times 15 \times 20 $  & \text{Conv:}~~$1 \times 1$ \\\cline{3-4}
				&   &	 $ b \times 640 \times 15 \times 20 $  &   \text{Average Pool : } $2 \times 2 $; Stride: $2 \times 2$  \\
				\hline
				
				\vspace{0.5cm}
				$12$ & \vspace{0.5cm} Dense Block - $4$ & \vspace{0.5cm} $ b \times 1664 \times 15 \times 20 $  &  \[ \left[  \frac{ \text{Conv:~~} 1 \times 1 } {\text{Conv:~~} 3 \times 3  } \right] \times 6\]   \\
				\hline
				$13$ & Batch Norm - $5$ & $ b \times 1664 \times 15 \times 20 $  &  $\epsilon = 10 ^{-5}$; momentum = $0.1$  \\
				\hline
				\hline
			\end{tabular}
			\label{MTL_accuracy_4}
		}
		\vspace{-0.5cm}
	\end{table}	
	
	For the decoder part, we pass the output of \emph{Batch Norm -5} layer (see row $13$ of Table.~\ref{MTL_accuracy_4}) through \emph{ReLU} activation, succeeded by the following convolutional layer (see Table.~\ref{MTL_accuracy_6}), where $\beta = F$ and $\zeta = F$ to generate a tensor of size $[ t1 : b \times F \times m \times n ]$; where $m = 15$, $n = 20$ and $F = 1664$ respectively.
	\begin{table}[h!]
		\small{
			\caption{ The convolution layer }
			\centering
			\begin{tabular}{|P{2.9cm}| P{9.1cm}|}
				\hline
				
				\vspace{0.01 cm}Convolution &	 No. of input filters/channels : $\beta$; No. of output filters/channels : $\zeta$; Conv kernel : $1 \times 1$; Stride: $1 \times 1$; Padding: $0$ \\
				\hline
			\end{tabular}
			\label{MTL_accuracy_6}
		}
		\vspace{-0.5cm}
	\end{table}
	Now, to concatenate the output tensor (\emph{Trans-2}) from \emph{Transition Layer-2} (see item $9$ in Table.\ref{MTL_accuracy_4}) i.e. the tensor of $[b \times 256 \times 30 \times 40]$ dimension with $t1$, we perform an up-sampling to double the size of $t1$ by using bi-linear interpolation which generates a tensor of size $[b \times F \times (m\times2) \times (n\times2)]$. Now the tensor $t1$ and \emph{Trans-2} are concatenated along it's $2^{nd}$ dimension to generate another tensor of $[b \times 1920 \times (m\times2) \times (n\times2)]$ dimension. Then this tensor is passed through \emph{Convolution-A} layer (see item $1$ of Table.~\ref{MTL_accuracy_5}) to generate a tensor of size  $[b \times 832 \times (m\times2) \times (n\times2)]$; where $\beta$ and $\zeta$ is taken as $1920$ and $\frac{F}{2}=832$ respectively. Then this generated tensor is sequentially passed through \emph{Leaky ReLU-A}, \emph{Convolution-B} and \emph{Leaky ReLU-B} layers (see items $2$, $3$ and $4$ in Table.~\ref{MTL_accuracy_5}) to generate the tensor of size $[t2 : b \times 832 \times (m\times2) \times (n\times2)]$; where $\beta = \zeta = 832$ are taken for \emph{Convolution-B} layer (see item $3$ in  Table.~\ref{MTL_accuracy_5}).  
	
	\begin{table}[h!]
		\small{
			\caption{ The up-sampling block }
			\centering
			\begin{tabular}{|P{1cm}||P{2.9cm}| P{7.7cm}|}
				\hline
				
				\textbf{Sl. No.} & \textbf{Layers} & \textbf{Specifications}   \\ \hline		 
				$1$ & Convolution-A &	 No. of input filters/channels : $\beta$; No. of output filters/channels : $\zeta$; Conv kernel : $3 \times 3$; Stride: $1$; Padding: $1$ \\
				\hline
				$2$ & Leaky ReLU-A &	$\alpha = 0.2$; where $\alpha$ controls the angle of the negative slope \\
				\hline
				$3$ & Convolution-B &	 No. of input filters/channels : $\beta$; No. of output filters/channels : $\zeta$; Conv kernel : $3 \times 3$; Stride: $1$; Padding: $1$ \\
				\hline
				$4$ & Leaky ReLU-B &	$\alpha = 0.2$ \\
				\hline
				\hline
			\end{tabular}
			\label{MTL_accuracy_5}
		}
	\end{table}	
	Now in the same manner, the tensor $t2$ is interpolated and concatenated with the output tensor (\emph{Trans-1}) from \emph{Transition Layer-1} (see item $7$ in Table.\ref{MTL_accuracy_4}) i.e. the tensor of $[b \times 128 \times 60 \times 80]$ dimension. The concatenated tensor become of size $[b \times 960 \times 60 \times 80]$. This concatenated tensor is similarly passed through \emph{Convolution-A}, \emph{Leaky ReLU-A}, \emph{Convolution-B} and \emph{Leaky ReLU-B} layers (see items $1$, $2$, $3$ and $4$ in Table.~\ref{MTL_accuracy_5}), where $\beta = \frac{F}{2} + 128$ and $\zeta = \frac{F}{4}$ for \emph{Convolution-A} layer and $\beta = \frac{F}{4}$ and $\zeta = \frac{F}{4}$ for \emph{Convolution-B} layer. From these operations, we will generate a tensor of $[t3:  b \times 416 \times 60 \times 80]$ dimension.
	
	Then, the tensor $t3$ is again interpolated and concatenated with the output tensor from \emph{Max Pooling Layer} (see item $5$ in Table.\ref{MTL_accuracy_4}) i.e. the tensor of $[b \times 64 \times 120 \times 160]$ dimension. The concatenated tensor become of size $[b \times 480 \times 120 \times 160]$. This concatenated tensor is similarly passed through \emph{Convolution-A}, \emph{Leaky ReLU-A}, \emph{Convolution-B} and \emph{Leaky ReLU-B} layers (see items $1$, $2$, $3$ and $4$ in Table.~\ref{MTL_accuracy_5}), where $\beta = \frac{F}{4} + 64$ and $\zeta = \frac{F}{8}$ for \emph{Convolution-A} layer and $\beta = \frac{F}{8}$ and $\zeta = \frac{F}{8}$ for \emph{Convolution-B} layer. From these operations, we will generate a tensor of $[t4:  b \times 208 \times 120 \times 160]$ dimension.
	
	After that, the tensor $t4$ is also interpolated and concatenated with the output tensor from \emph{ReLU activation Layer} (see item $4$ in Table.\ref{MTL_accuracy_4}) i.e. the tensor of $[b \times 64 \times 240 \times 320]$ dimension. The concatenated tensor become of size $[b \times 272 \times 240 \times 320]$. This concatenated tensor is similarly pass through \emph{Convolution-A}, \emph{Leaky ReLU-A}, \emph{Convolution-B} and \emph{Leaky ReLU-B} layers (see items $1$, $2$, $3$ and $4$ in Table.~\ref{MTL_accuracy_5}), where $\beta = \frac{F}{8} + 64$ and $\zeta = \frac{F}{16}$ for \emph{Convolution-A} layer and $\beta = \frac{F}{16}$ and $\zeta = \frac{F}{16}$ for \emph{Convolution-B} layer. From these operations, we will generate a tensor of $[t5:  b \times 104 \times 240 \times 320]$ dimension. Finally, this $t5$ tensor is passed through the following convolution layer where $\beta = \frac{F}{16}$ and $\zeta = 3$ to generate a tensor of $[t6:  b \times 3 \times 240 \times 320]$ dimension.
	\begin{table}[]
		\small{
			\centering
			\begin{tabular}{|P{2.9cm}| P{9.1cm}|}
				\hline
				
				\vspace{0.01 cm} \textbf{Convolution} &	 No. of input filters/channels : $\beta$; No. of output filters/channels : $\zeta$; Conv kernel : $3 \times 3$; Stride: $1 \times 1$; Padding: $1$ \\
				\hline
			\end{tabular}
			\label{MTL_accuracy_8}
		}
		\vspace{-0.5cm}
	\end{table}
	The input images are represented by their original colors in the range $[0,1]$ without any input data normalization. Target depth maps are clipped to the range $[0.4, 10]$ in meters.

	\section{Further Details : Learning Loss Function}
	\emph{Technique-1} is explained in details in the main paper. In this section of supplementary materials, we will further describe \emph{Technique-2} and its variants i.e. \emph{Variant-1} and \emph{Variant-2} of \emph{Technique-2} in details.
	
	\subsection{Technique 1} As the first approach, the total loss ($L_{total}$) is computed by adding the loss over \emph{depth} image i.e. $L_d$ and the loss over \emph{estimated simulated image} i.e. $L_p$. Hence, $L_{total} = L_d + L_p$. Where, $L_d$ is computed by combining the weights $\lambda_1^y$ and $\lambda_2^y$ (see Equation~\eqref{depth_equation}) and $L_p$ is computed by combining the weights $\lambda_1^q$ and $\lambda_2^q$ (see Equation~\ref{predicted_equation}). These weights are initially set as $0.1$.
	
	\subsubsection{Variant 1 :} 
	\label{tech_1_variant_1_weight}
	Instead of fixing the values of  $\lambda_1^q$ and $\lambda_2^q$ as $0.1$, we compute them automatically by using the same network, shown in Fig.~\ref{fig:archi_model}. The pre-trained features (of size $[b \times 1664 \times 15 \times 20]$ from the last layer of the \emph{DenseNet} based encoder i.e. from \emph{Batch Norm - 5} layer (see item $13$ in Table.~\ref{MTL_accuracy_3_1}) is branched out and passed through two consecutive \emph{ConvBNRelu} blocks, shown in following Table.~\ref{MTL_accuracy_9}; where $\beta = 1664$, $\zeta = 512$, $k = 11$, $s = 1$ and $p = 1$ is taken for first \emph{ConvBNRelu} block and $\beta = 512$, $\zeta = 256$, $k = 9$, $s = 1$ and $p = 1$ for the second \emph{ConvBNRelu} block. Then the output feature is passed through the ``Average Pooling'' layer\footnote{here we have used ``Adaptive Average Pooling'' algorithm from PyTorch library. For more details, see : \url{https://pytorch.org/cppdocs/api/classtorch_1_1nn_1_1_adaptive_avg_pool1d.html} }, having a kernel of size $1$. This makes the feature to get transformed into $2D$ features which are then flattened and reduced into the $1D$ feature. Then these $1D$ features are passed through following layers:
	\vspace{2mm}
	\fbox{\parbox{\dimexpr\linewidth-0\fboxsep-2\fboxrule\relax}{\centering 
			\strut \textbf{DropOut}-\textbf{FC}($256 \rightarrow 128$)-\textbf{RELU}
	}} 
	\vspace{2mm}
	\fbox{\parbox{\dimexpr\linewidth-0\fboxsep-2\fboxrule\relax}{\centering 
			\strut \textbf{DropOut}-\textbf{FC}($128 \rightarrow 64$)-\textbf{RELU}
	}}  
	\vspace{2mm}
	\fbox{\parbox{\dimexpr\linewidth-0\fboxsep-2\fboxrule\relax}{\centering 
			\textbf{DropOut}-\textbf{FC}($64 \rightarrow 32$)-\textbf{RELU}
	}}  
	\vspace{2mm}
	\fbox{\parbox{\dimexpr\linewidth-0\fboxsep-2\fboxrule\relax}{\centering 
			\textbf{DropOut}-\textbf{FC}($32 \rightarrow 16$)-\textbf{RELU}
	}}
	\vspace{2mm}
	\fbox{\parbox{\dimexpr\linewidth-0\fboxsep-2\fboxrule\relax}{\centering 
			\textbf{FC}($16 \rightarrow 2$)
	}}
	
	After passing through the above layers, we will obtain two output values which are then passed through \emph{Sigmoid} activation function to finally get two weight values. We apply this above mentioned technique to obtain two weight values i.e. $w^1_{Depth}$ and $w^2_{Depth}$ from the \emph{encoder} of \emph{Block-1} and another two weight values i.e. $w^1_{Residue}$ and $w^2_{Residue}$ from the encoder of \emph{Block-2} (see Fig.~\ref{fig:archi_model} for reference).
	
	\begin{table}[h!]
		\small{
			\caption{ The \emph{ConvBNRelu} block }
			\centering
			\begin{tabular}{|P{1cm}||P{2.9cm}| P{7.7cm}|}
				\hline
				\textbf{Sl. No.} & \textbf{Layers} & \textbf{Specifications}   \\ \hline		 
				$1$ & Convolution&	 No. of input filters/channels : $\beta$; No. of output filters/channels : $\zeta$; Conv kernel : $k \times k$; Stride: $s$; Padding: $p$ \\
				\hline
				$2$ & Batch Norm &	No. of output filters/channels : $\zeta$ \\
				\hline
				$2$ & ReLU & \\
				\hline
				\hline
			\end{tabular}
			\label{MTL_accuracy_9}
		}
	\end{table}
	
	Please note that here we have applied the \emph{Sigmoid} activation function because the posterior probability values are between $0-1$ but the sum of these values can be greater than $1$ (whereas, in the case of \emph{SoftMax} activation function, the posterior probability values are between $0-1$ and the sum of all these values are $1$)\footnote{For details, see : \footnotesize{https://medium.com/arteos-ai/the-differences-between-sigmoid-and-softmax-activation-function-12adee8cf322}}. Hence, by applying \emph{Sigmoid} activation function, we confirm that the individual weights i.e. $w^1_{Depth}$, $w^2_{Depth}$, $w^1_{Residue}$ and $w^2_{Residue}$ are within $0-1$. Moreover, by applying $(1-w^1_{Depth})$ or $(1-w^1_{Residue})$ on the second term for loss computation (see Equation~\ref{tech_1_variant_1_equation_ver_1}), we confirm that only the remaining weight is applied on the second term. By using these automatic  weight values, the $L_d$ and $L_p$ is calculated as:
	
	\begin{equation}
		\begin{array}{l}
			L_d(y_p, \hat{y}_p) = w^1_{Depth} L_{depth}(y_p, \hat{y}_p) + (1- {w^1_{Depth}}) L_{SSIM}(y_p, \hat{y}_p) \\
			L_p(q_p, \hat{q}_p) = {w^1_{Residue}} ~\left[\frac{1}{n} \sum_{p}^{n}|q_p - \hat{q}_p|\right] + (1-{w^1_{Residue}}) ~\left[ \frac{1-SSIM(q_p, \hat{q}_p)}{2} \right]
			\\
			L_{total} = L_d(y_p, \hat{y}_p) + L_p(q_p, \hat{q}_p)
		\end{array}
		\label{tech_1_variant_1_equation_ver_1}
	\end{equation}
	
	\subsubsection{Variant 2 :} In addition with the weighted computation of $L_d(y_p, \hat{y}_p)$ and $L_p(q_p, \hat{q}_p)$ (according to Equation~\eqref{tech_1_variant_1_equation}), we also compute the total loss in the following manner, where the weight values $w^2_{Depth}$ and $(1 - {w^2_{Depth}})$ are applied to perform weighted combination to the computation of total loss (i.e. $L_{total}$). 
	\begin{equation}
		\begin{array}{l}
		L_{total} = {w^2_{Depth}} L_d(y_p, \hat{y}_p) + (1 - {w^2_{Depth}}) L_p(q_p, \hat{q}_p)
		\end{array}
		\label{tech_1_variant_2_equation_ver_1}
	\end{equation}
	
	\subsection{Technique 2} As a second approach, a loss term corresponding to the ``initial degraded'' image (computed by using Equation~\eqref{eq_1} in the main paper) is added, compared to the total loss of Equation~\eqref{tech_1_variant_1_equation}.
	
	\subsubsection{Variant 1 :} In the same way as it is mentioned in Equation~\eqref{tech_1_variant_1_equation}, here also we compute the weighted (automatic) version of $L_d(y, \hat{y})$ and $L_p(q, \hat{q})$. Whereas the weighted (automatic) version of $ L_t(h, \hat{h})$ and the total loss is computed in the following manner. The needed weight $w^2_{depth}$ is computed in the above defined manner.
	\begin{equation}
	\begin{array}{l}
	L_t(h_p, \hat{h}_p) = {w^2_{Depth}} ~\left[\frac{1}{n} \sum_{p}^{n}|h_p - \hat{h}_p|\right] + (1-{w^2_{Depth}}) ~\left[ \frac{1-SSIM(h_p, \hat{h}_p)}{2} \right]
	\\
	L_{total} = L_d(y_p, \hat{y}_p) + L_p(q_p, \hat{q}_p) + L_t(h_p, \hat{h}_p)
	\end{array}
	\label{total_loss_equation_tech_2_variant_1_ver1}
	\end{equation}
	
	\subsubsection{Variant 2 :}
	\label{tech_2_variant_2} 
	In the same manner, to compute the total loss, we need at-least $2$ weight values. To obtain these weights, we use the same strategy as the one mentioned in Section~\ref{tech_1_variant_1_weight}. We take out another branch from the last layer of the \emph{DenseNet} based encoder i.e. from \emph{Batch Norm - 5} layer (see item $13$ in Table.~\ref{MTL_accuracy_3_1}) and apply exactly same operation as before like passing through two consecutive \emph{ConvBNRelu} blocks, shown in following Table.~\ref{MTL_accuracy_9}, applying ``Adaptive Average Pooling'' layer, followed by flattening operation which is followed by several blocks of \fbox{ \strut \textbf{DropOut}-\textbf{FC}-\textbf{RELU}}. But here, from the very last linear layer (i.e. \fbox{ \strut \textbf{FC}($16 \rightarrow 3$ )}), we obtain $3$ output values which are then passed through \emph{Soft-Max} activation function to finally get $3$ weight values. 
	Please note that here we have applied the \emph{Soft-Max} activation function because the posterior probability values are between $0-1$ and the sum of all these values are $1$. Hence, by applying \emph{Sigmoid} activation function, we confirm that the individual weights i.e. $w^1_{Extra}$, $w^2_{Extra}$ are within $0-1$. Moreover, by applying $[1-({w^1_{Extra}} + {w^2_{Extra}})]$ on the third term for loss computation (see Equation~\ref{tech_2_variant_2_equation_ver_1}), we confirm that only the remaining weight is applied on the third term. 
	
	\begin{dmath}
		L_{total} = {w^1_{Extra}} \times L_d(y_p, \hat{y}_p) + {w^2_{Extra}} \times L_p(q_p, \hat{q}_p)  +  [1-({w^1_{Extra}} + {w^2_{Extra}})] \times L_t(h_p, \hat{h}_p)
		\label{tech_2_variant_2_equation_ver_1}
	\end{dmath}

	\subsection{Technique 3} As the third modification, here we introduce an additional \emph{encoder-decoder} block to directly estimate the simulated image $(I^{Direct}_{Predicted})$ from the clean \emph{RGB} image. Hence, we compute a dedicated loss ($L_g$) for $(I^{Direct}_{Predicted})$ image only:
	\begin{equation}
	L_g(s_p, \hat{s}_p) = \lambda_1^s ~\left[\frac{1}{n} \sum_{p}^{n}|s_p - \hat{s}_p|\right] + \lambda_2^s ~\left[ \frac{1-SSIM(s_p, \hat{s}_p)}{2} \right]
	\label{direct_loss_ver_1}
	\end{equation}
	where $s_p$ and $\hat{s}_p$ represents the true and predicted ``directly simulated'' image $({I}_{Initial}^{Degraded})$ by the network and the value of $\lambda_1^s$ and $\lambda_2^s$ are set to $0.1$. Finally, the total loss is: 
	\begin{equation}
	L_{total} = L_d(y_p, \hat{y}_p) + L_p(q_p, \hat{q}_p) + L_t(h_p, \hat{h}_p) + L_g(s_p, \hat{s}_p)
	\label{total_loss_equation_tech_3_ver_1}
	\end{equation}
	
	\subsubsection{Variant 1 :} As the \emph{variant 1} under this category, in the same way as it is mentioned in Equation~\ref{total_loss_equation_tech_2_variant_1}, here also we compute the weighted (automatic) version of  $L_g(s_p, \hat{s}_p)$ in addition to the weighted version of $L_d(y_p, \hat{y}_p)$, $L_p(q_p, \hat{q}_p)$ and $L_t(h_p, \hat{h}_p)$ (see Equation~\eqref{total_loss_equation_tech_3_variant_1_ver_1}) in the following manner. Here also the weight value of $w^1_{Direct}$ is applied in the same manner to compute the weighted combination of total loss $(L_{total})$.  
	\begin{equation}
	\begin{array}{l}
	L_g(s_p, \hat{s}_p) = {w^1_{Direct}} ~\left[\frac{1}{n} \sum_{p}^{n}|s_p - \hat{s}_p|\right] + (1-{w^1_{Direct}}) ~\left[ \frac{1-SSIM(s_p, \hat{s}_p)}{2} \right]
	\\
	L_{total} = L_d(y_p, \hat{y}_p) + L_p(q_p, \hat{q}_p) + L_t(h_p, \hat{h}_p) + L_g(s_p, \hat{s}_p)
	\end{array}
	\label{total_loss_equation_tech_3_variant_1_ver_1}
	\end{equation}
	
	\paragraph{Variant 2 :} As the \emph{variant 2} under this category, in the same manner, here also we can compute the total loss in the following way. In this case, we need at-least $3$ weight values which are obtained in the same manner as it is mentioned in Section~\ref{tech_2_variant_2}. 	 
	
	\begin{dmath}
		L_{total} = {w^1_{Extra}} \times L_d(y_p, \hat{y}_p) + {w^2_{Extra}} \times L_p(q_p, \hat{q}_p) + {w^3_{Extra}} \times L_t(h_p, \hat{h}_p) +  [1-({w^1_{Extra}} + {w^2_{Extra}} + {w^3_{Extra}} )] \times L_t(h_p, \hat{h}_p)
		\label{tech_3_variant_2_equation_ver_1}
	\end{dmath}
	One important thing to note here is that all the computed weight values $(\mathcal{W})$ e.g. $ \mathcal{W}: w^1_{depth}$,  $w^2_{depth}$,  $w^1_{Residue}$, $w^2_{Residue}$ etc. computes weights based on a given RGB image. Hence, if there are $b$ number of images in a batch then we will generate $b$ number of such weights. But all these Equations for computing loss e.g. Equation~\eqref{tech_1_variant_1_equation_ver_1}, \eqref{tech_1_variant_2_equation_ver_1}, \eqref{total_loss_equation_tech_2_variant_1_ver1}, \eqref{tech_2_variant_2_equation_ver_1} etc. are directly computed on batches. Hence, to apply the weight values, we take it's mean over a batch i.e. $w = \frac{\sum_{n = 1}^{b} \mathcal{W}_n}{b}$. This is also a reason to apply $1-w$ amount of weights in the second term of the loss calculation equations e.g. Equation~\eqref{tech_1_variant_1_equation_ver_1}, \eqref{tech_1_variant_2_equation_ver_1} etc.

	\begin{table}[]
		\small{
			\caption{The number of trainable parameters of each block for all the proposed techniques. The time needed to execute each epoch for each of the techniques. The extra parameters in \emph{Variant-1} and \emph{Variant-2} of each block appears due to supplementary branch, needed for the automatic weight computation. }
			\centering
			\begin{tabular}{|P{2.9cm}|P{2.0cm} P{2.0cm} P{2.0cm} || P{2.6cm}|}
				\hline
				
				\textbf{Proposed Method} & \textbf{Block-1} & \textbf{Block-2} & \textbf{Block-3}  & \textbf{Training Time}   \\ \hline
				
				\mycc Technique-1 &	\mycc $443,22,689$ &  \mycc $443,24,563$ & \mycc  $ \times$ & 2 hr 39 min 25 sec\\
				\hline
				\mycc Technique-1 Variant-1 & \mycc $1580,73,747$ &  \mycc $1580,75,621$ & \mycc $\times$ & 3 hr 21 min 35 sec  \\
				\hline
				\mycc Technique-1 Variant-2 & \mycc $1580,73,747$ &  \mycc $1580,75,621$ & \mycc $\times$ & 3 hr 28 min 19 sec \\
				\hline
				\hline
				\mybb Technique-2 &	\mybb $443,22,689$ &  \mybb $443,22,689$ & \mybb  $ \times$ & 2 hr 26 min 56 sec \\
				\hline
				\mybb Technique-2 Variant-1 & \mybb $1580,73,747$ &  \mybb $1580,75,621$ & \mybb $\times$ & 2 hr 49 min 17 sec  \\
				\hline
				\mybb Technique-2 Variant-2 & \mybb $1580,73,747$ &  \mybb $1580,75,621$ & \mybb $\times$ & 2 hr 49 min 18 sec \\
				\hline
				\hline
				\myrr Technique-3 &	\myrr $443,22,689$ &  \myrr $443,24,563$ & \myrr  $ 443,24,563 $ & 5 hr 41 min 59 sec \\
				\hline
				\myrr Technique-3 Variant-1 & \myrr $1580,73,747$ &  \myrr $1580,75,621$ & \myrr $ 1580,75,621 $ & 5 hr 59 min 24 sec  \\
				\hline
				\myrr Technique-3 Variant-2 & \myrr $1581,17,558$ &  \myrr $1580,75,621$ & \myrr $1580,75,621$  & 5 hr 59 min 27 sec \\
				\hline
				\hline
			\end{tabular}
			\label{MTL_accuracy_3_1}
		}
		\vspace{-0.5cm}
	\end{table}	
	\section{Training Details}
	For each of the different proposed network configurations, the total number of trainable parameters are mentioned in Table.~\ref {MTL_accuracy_3_1}. We have also mentioned the time needed to execute one epoch for each of the proposed techniques. It can be seen that \emph{Technique-3} and two of its variants have a lot more training parameters and it also takes more time to train. The training parameters are mentioned in Table.~\ref{MTL_accuracy_4_1}. We have used \emph{Adam Optimizer} from PyTorch\footnote{\url{https://pytorch.org/docs/stable/generated/torch.optim.Adam.html}} library for the training. Except learning rate, we have taken the default values of other parameters. 
	\begin{table}[]
		\small{
			\caption{The parameters for training}
			\begin{center}
				\begin{tabular}{||P{2.9cm}|P{2.0cm} P{2.0cm}||}
					\hline
					
					\textbf{Proposed Method} & \textbf{Optimizer} & \textbf{Learning Rate}  \\ \hline
					
					\mycc Technique-1 &	\mycc Adam &  \mycc $1e^{-6}$ \\
					\hline
					\mycc Technique-1 Variant-1 & \mycc Adam &  \mycc $1e^{-6}$\\
					\hline
					\mycc Technique-1 Variant-2 & \mycc Adam &  \mycc $1e^{-6}$  \\
					\hline
					\hline
					\mybb Technique-2 &	\mybb Adam &  \mybb $1e^{-6}$ \\
					\hline
					\mybb Technique-2 Variant-1 & \mybb Adam &  \mybb $1e^{-6}$ \\
					\hline
					\mybb Technique-2 Variant-2 & \mybb Adam &  \mybb $1e^{-6}$ \\
					\hline
					\hline
					\myrr Technique-3 &	\myrr Adam &  \myrr $1e^{-5}$\\
					\hline
					\myrr Technique-3 Variant-1 & \myrr Adam &  \myrr $1e^{-5}$\\
					\hline
					\myrr Technique-3 Variant-2 & \myrr Adam &  \myrr $1e^{-6}$ \\
					\hline
					\hline
				\end{tabular}
			\end{center}
			\label{MTL_accuracy_4_1}
		}
		\vspace{-0.5cm}
	\end{table}	
	
	\begin{figure}[!h]	       
		\begin{subfigure}[b]{\linewidth}
			\centering{\includegraphics[trim = 5.2cm 1.6cm 4.1cm 1.6cm, clip,  scale = 0.30]{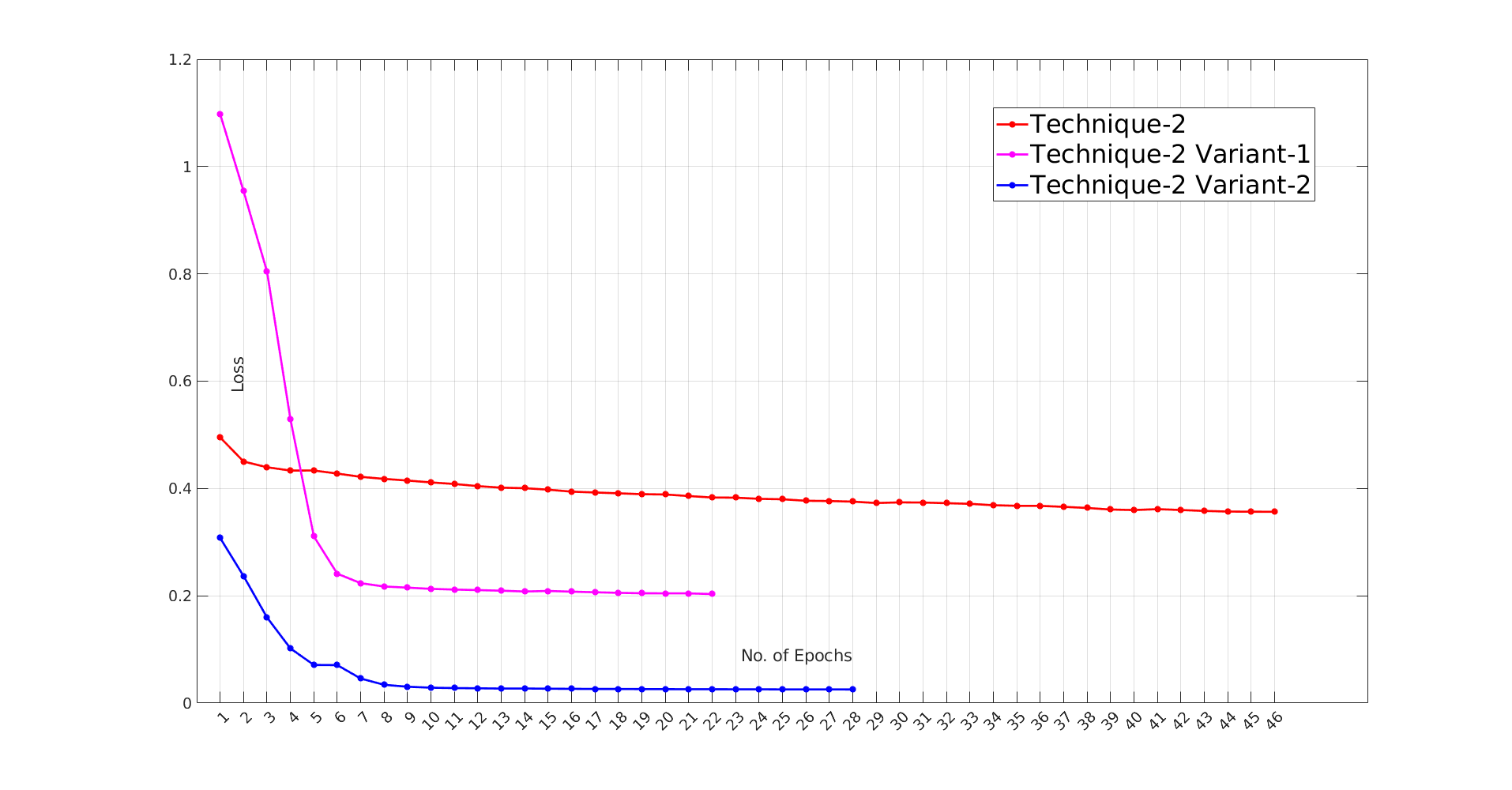}}
			\caption{}
			\label{fig:archi_water_1}
		\end{subfigure}
		
		\begin{subfigure}[b]{\linewidth}
			\centering{\includegraphics[trim = 5.2cm 1.6cm 4.1cm 1.6cm, clip,  scale = 0.30]{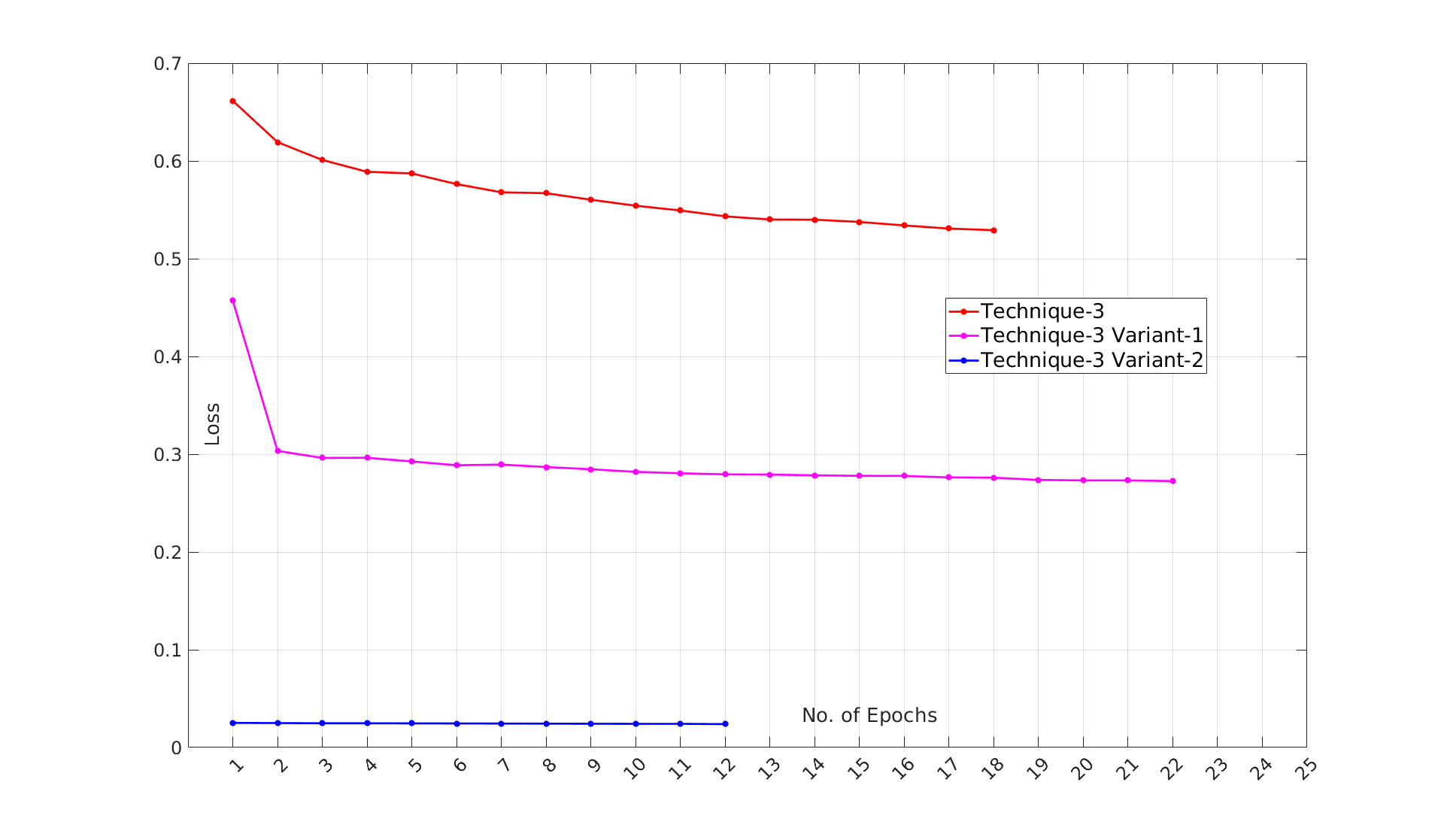}}
			\caption{}
			\label{fig:archi_water_2}
		\end{subfigure}	
		\caption{\small{(a) The training loss curve of \emph{Technique-2} and it's variants  (b) The training loss curve of \emph{Technique-3} and it's variants } 
		}
		\label{fig:archi_water}
	\end{figure}
	
	\subsection{Training Loss}
	The training loss plots of the \emph{Technique-2}, \emph{Technique-2 Variant-1}, \emph{Technique-2 Variant-2} and consecutively the training loss of \emph{Technique-3}, \emph{Technique-3 Variant-1}, \emph{Technique-3 Variant-2} are shown in Fig.~\ref{fig:archi_water_1} and Fig.~\ref{fig:archi_water_2} respectively. It can be visible from these plots that the training losses of all the techniques decreases gradually whereas the \emph{Variant-1} and \emph{Variant-2} of both the \emph{Technique-2} and \emph{Technique-3} gets stabilized quickly (i.e. less number of epochs are needed). Moreover, we can also see that either training loss decreases rapidly (see the plots of \emph{Technique-2 Variant-1}, \emph{Technique-2 Variant-2} and \emph{Technique-3 Variant-1}) or it starts with low value and stabilizes quickly (see the plot of \emph{Technique-3 Variant-1}). Please note that \emph{Technique-1} along with it's variants has similar characteristics and performance as \emph{Technique-2} and it's variants, here we have shown the loss curve on \emph{Technique-2} and it's variants only. 
	
	\begin{figure*}[bht!]
		\centering	
		\begin{subfigure}[b]{\textwidth}
			\centering {\includegraphics[trim = 0.0cm 8.6cm 0.0cm 0.0cm, clip, scale = 0.18]{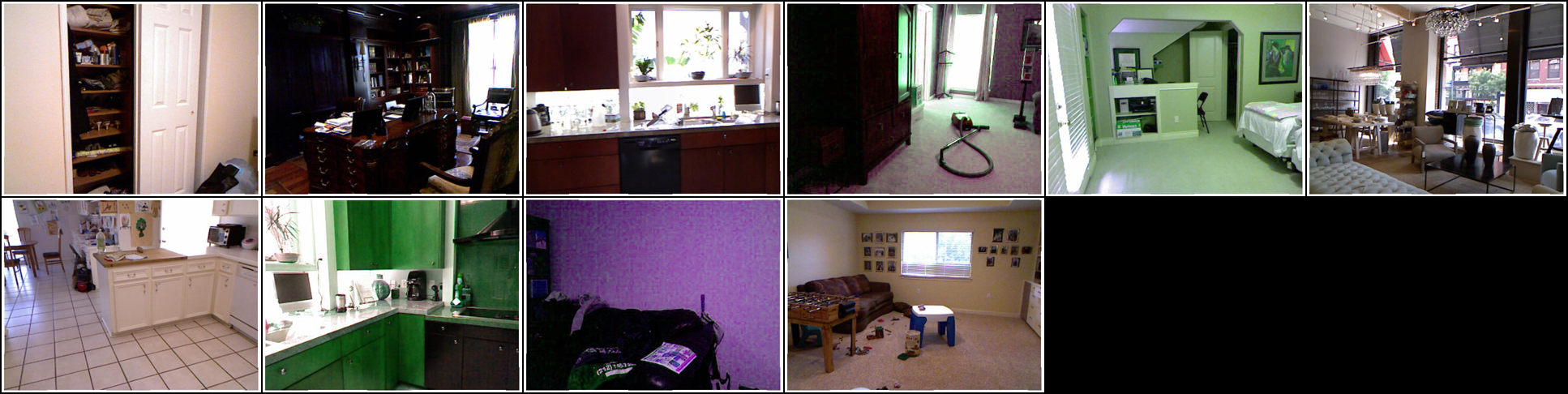} }
			\centering
			\vspace{-5mm}
			\caption{}
			\label{fig:technique_0_orig_half}
		\end{subfigure}			
		\begin{subfigure}[b]{\textwidth}
			\centering {\includegraphics[trim = 0.0cm 8.6cm 0.0cm 0.0cm, clip, scale = 0.18]{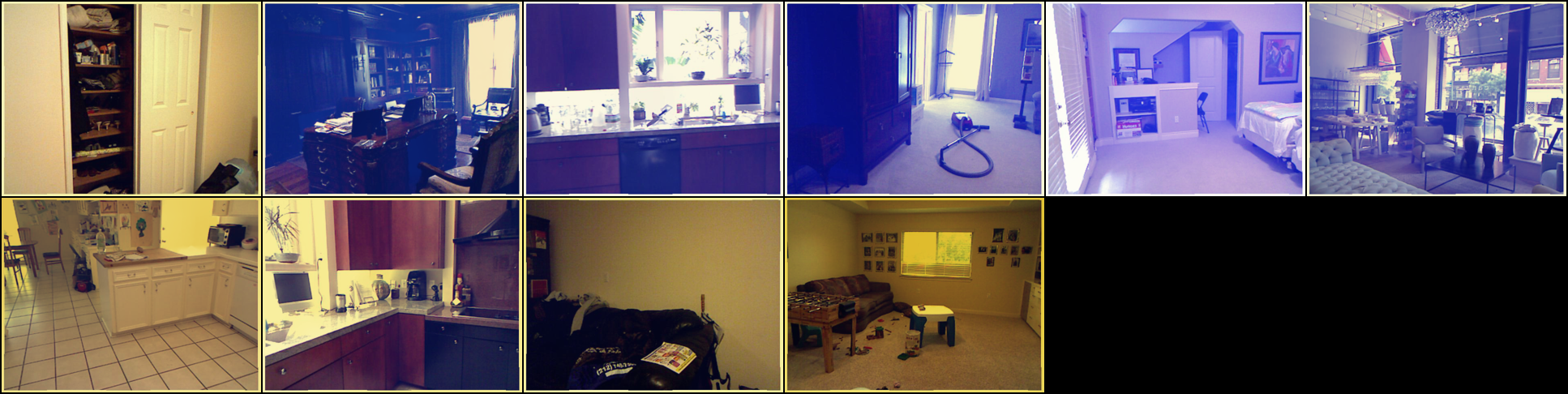} }
			\centering
			\vspace{-5mm}
			\caption{}
			\label{fig:technique_0_haze_GT_half}
		\end{subfigure} 
		\begin{subfigure}[b]{\textwidth}
			\centering {\includegraphics[trim = 0.0cm 8.6cm 0.0cm 0.0cm, clip, scale = 0.18]{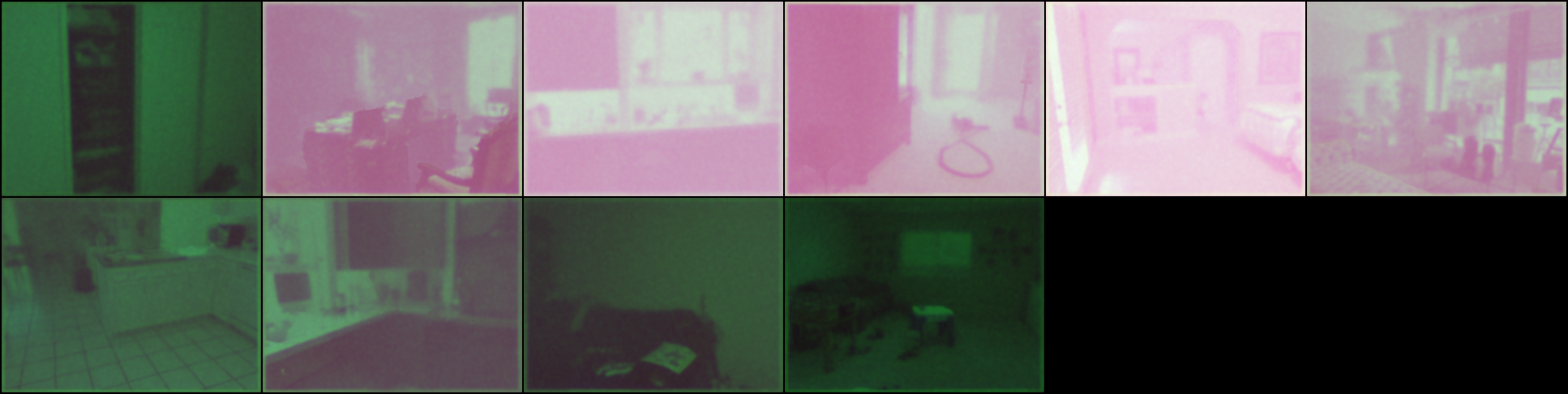} }
			\centering
			\vspace{-5mm}
			\caption{}
			\label{fig:technique_0_complex_GT_half}
		\end{subfigure}
		\begin{subfigure}[b]{\textwidth}
			\centering {\includegraphics[trim = 0.0cm 8.6cm 0.0cm 0.0cm, clip, scale = 0.18]{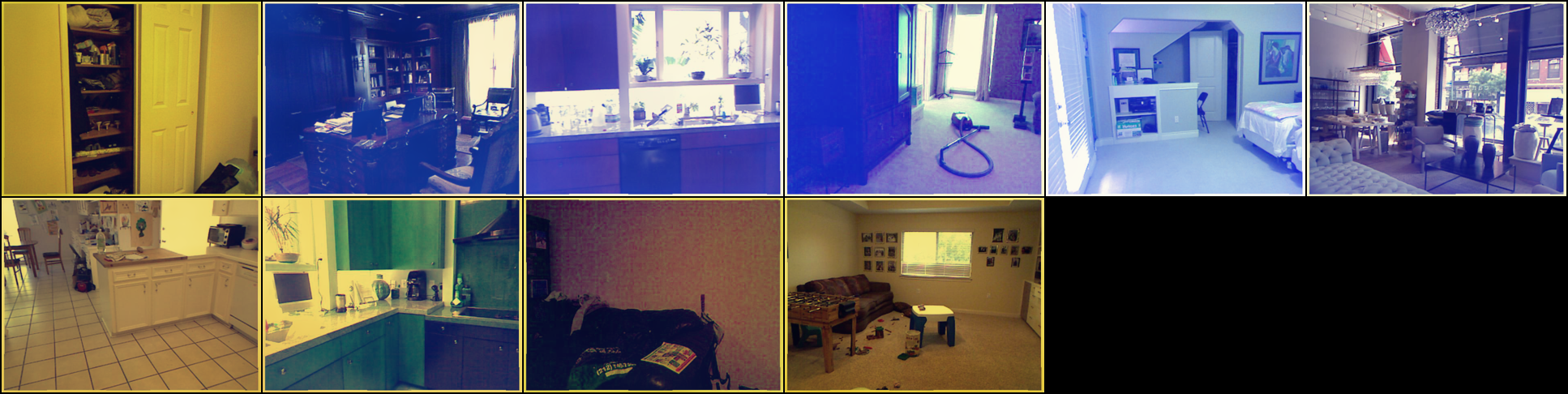} }
			\centering
			\vspace{-5mm}
			\caption{}
			\label{fig:technique_0_haze_Pred_half}
		\end{subfigure}
		\begin{subfigure}[b]{\textwidth}
			\centering {\includegraphics[trim = 0.0cm 8.6cm 0.0cm 0.0cm, clip, scale = 0.18]{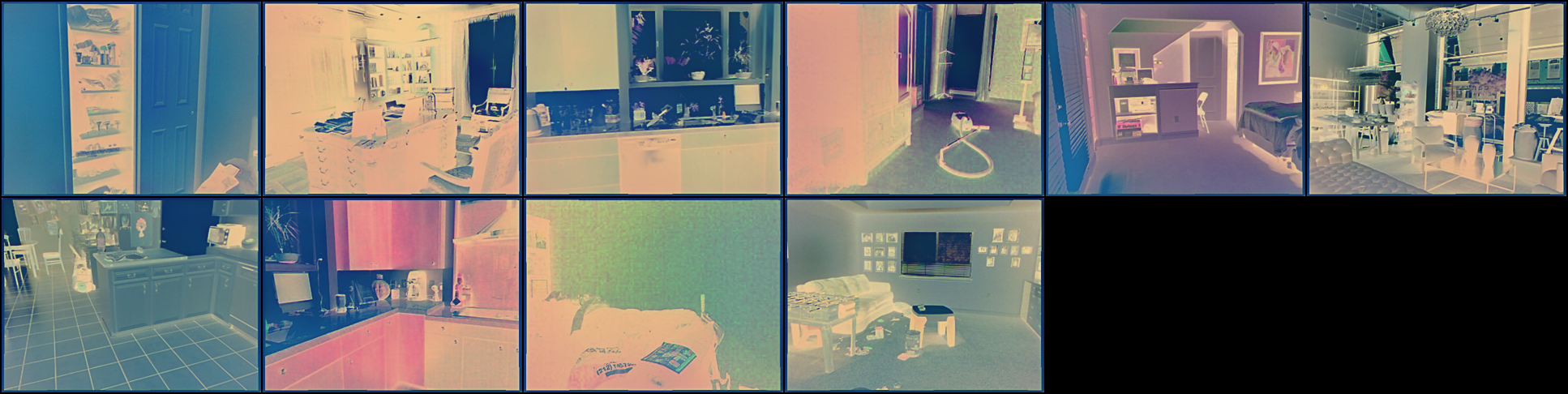} }
			\centering
			\vspace{-5mm}
			\caption{}
			\label{fig:technique_0_Residue_Pred_half}
		\end{subfigure}
		\begin{subfigure}[b]{\textwidth}
			\centering {\includegraphics[trim = 0.0cm 8.6cm 0.0cm 0.0cm, clip, scale = 0.18]{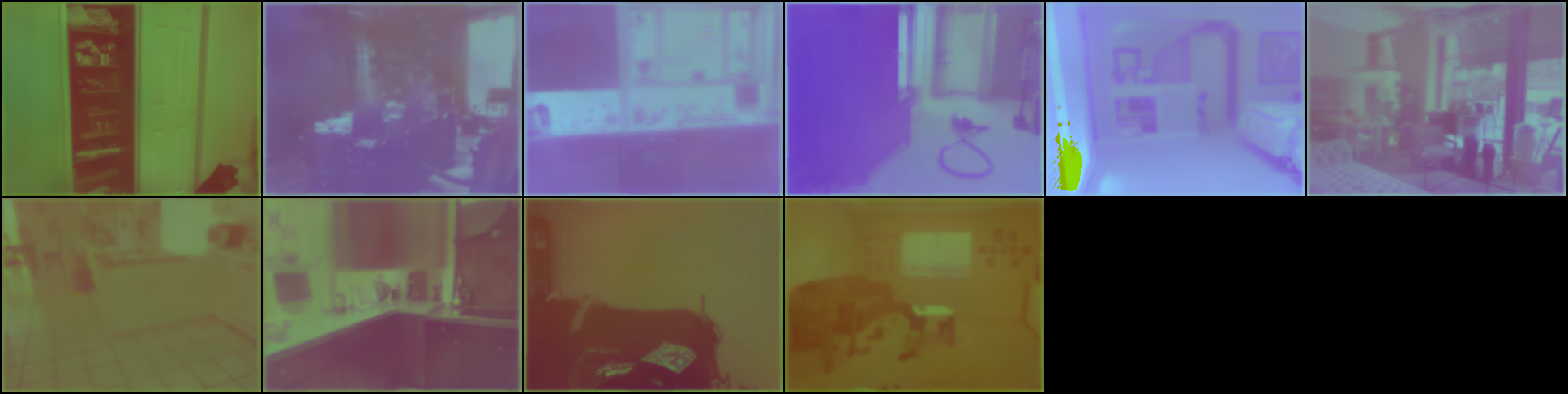} }
			\centering
			\vspace{-5mm}
			\caption{}
			\label{fig:technique_0_Complex_Pred_half}
		\end{subfigure}
		\caption[endProcess]{ \textbf{Qualitative Measures obtained by \textit{Proposed} Technique 1 :} (a) Original \emph{RGB} images (i.e. $J_c(\mathbf{x})$) (b) Ground Truth of ``Initial Degraded Image'' (i.e. $I_c(\mathbf{x})$) (c)  Ground Truth of ``Simulated Underwater Image'' (d) Predicted ``Initial Degraded Image'' (i.e. $I^{Degraded}_{Initial}$) (e) Predicted ``Residual Image'' (i.e. $I^{Residue}$) (f) Predicted ``Simulated Underwater Image'' (i.e. $\hat{I}^{Simulated}_{Predicted}$).  }
		\label{fig:all_single_task_0}
		\vspace{-0.5cm}
	\end{figure*}

	\begin{figure*}[bht!]
		\centering	
		\begin{subfigure}[b]{\textwidth}
			\centering {\includegraphics[trim = 0.0cm 8.6cm 0.0cm 0.0cm, clip, scale = 0.18]{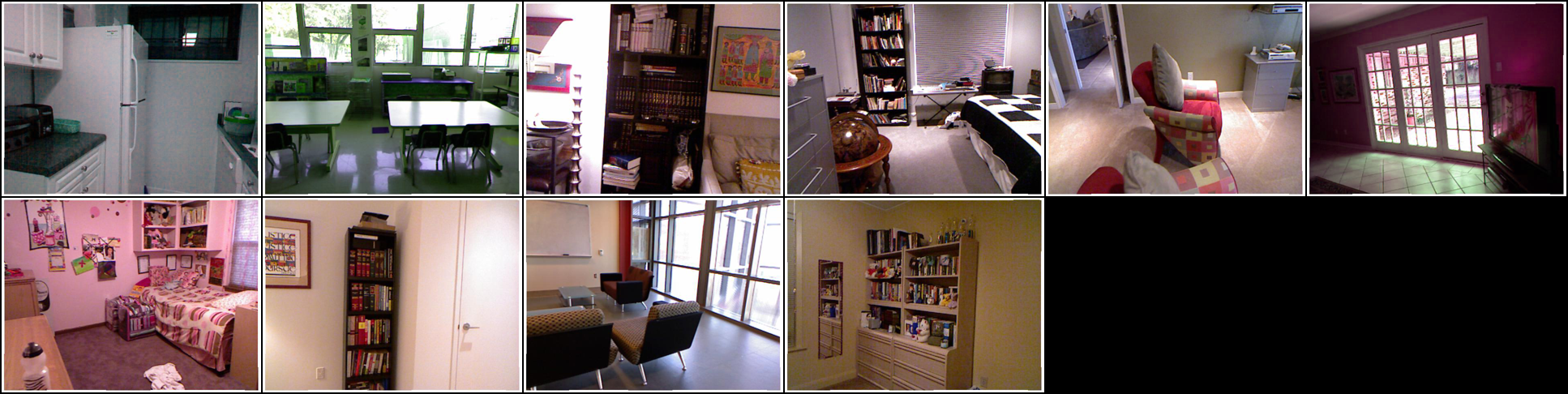} }
			\centering
			\vspace{-5mm}
			\caption{}
			\label{fig:technique_1_var_1_orig_half}
		\end{subfigure}			
		\begin{subfigure}[b]{\textwidth}
			\centering {\includegraphics[trim = 0.0cm 8.6cm 0.0cm 0.0cm, clip, scale = 0.18]{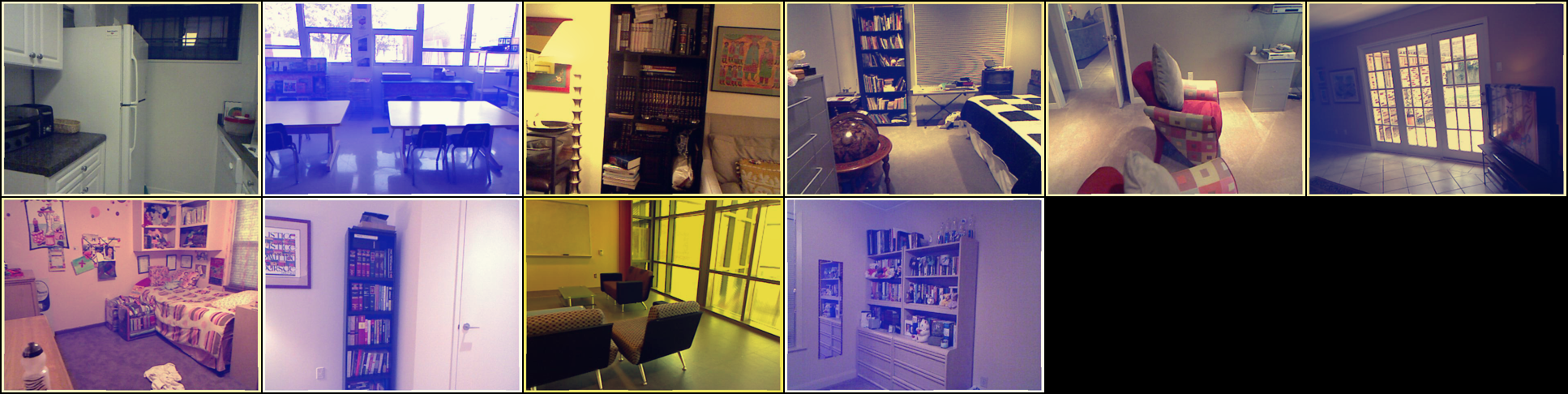} }
			\centering
			\vspace{-5mm}
			\caption{}
			\label{fig:technique_1_var_1_haze_GT_half}
		\end{subfigure} 
		\begin{subfigure}[b]{\textwidth}
			\centering {\includegraphics[trim = 0.0cm 8.6cm 0.0cm 0.0cm, clip, scale = 0.18]{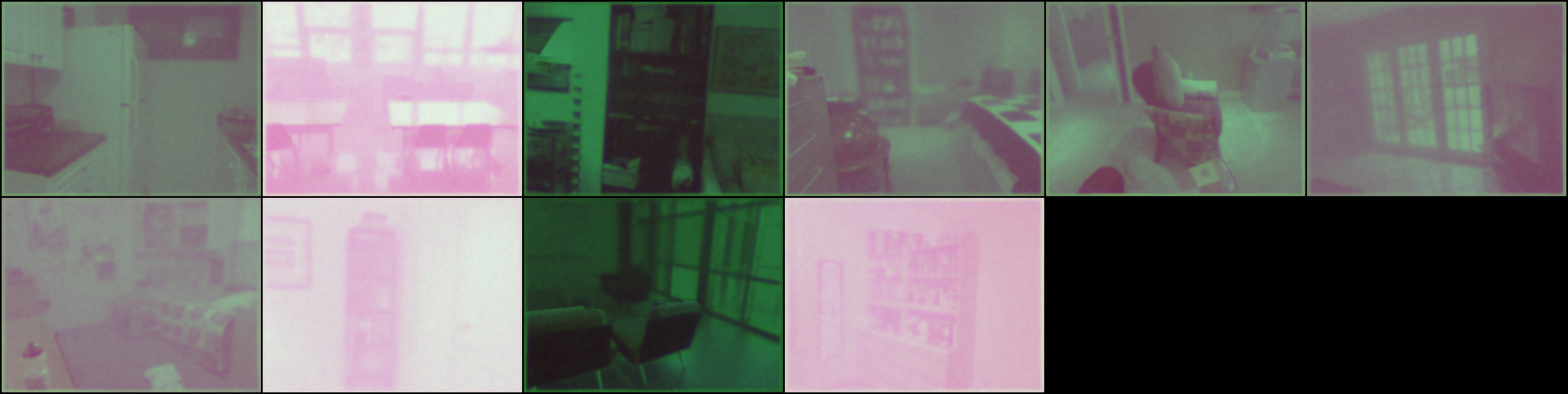} }
			\centering
			\vspace{-5mm}
			\caption{}
			\label{fig:technique_1_var_1_complex_GT_half}
		\end{subfigure}
		\begin{subfigure}[b]{\textwidth}
			\centering {\includegraphics[trim = 0.0cm 8.6cm 0.0cm 0.0cm, clip, scale = 0.18]{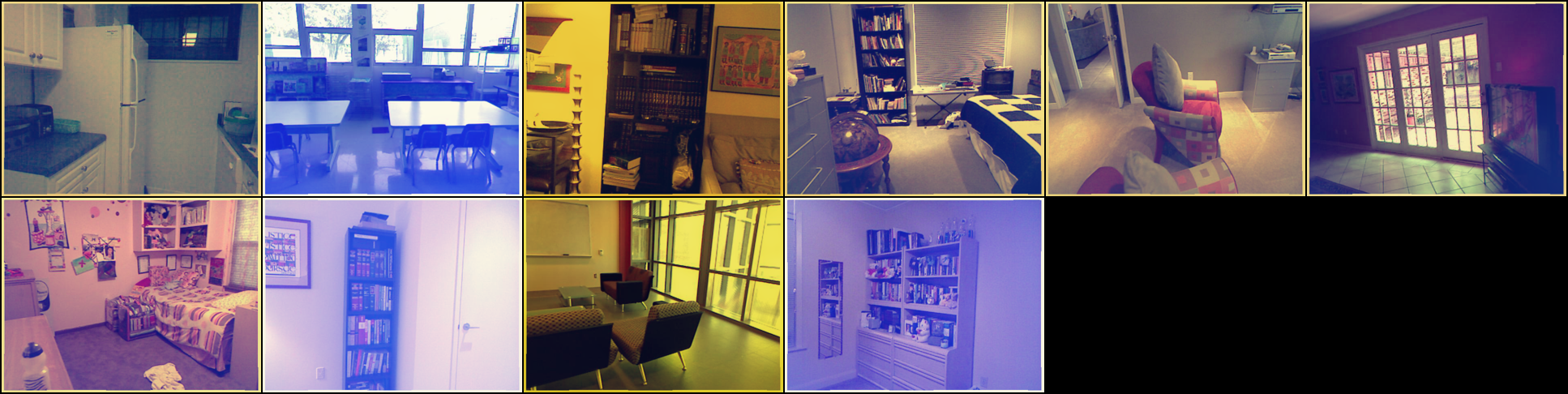} }
			\centering
			\vspace{-5mm}
			\caption{}
			\label{fig:technique_1_var_1_haze_Pred_half}
		\end{subfigure}
		\begin{subfigure}[b]{\textwidth}
			\centering {\includegraphics[trim = 0.0cm 8.6cm 0.0cm 0.0cm, clip, scale = 0.18]{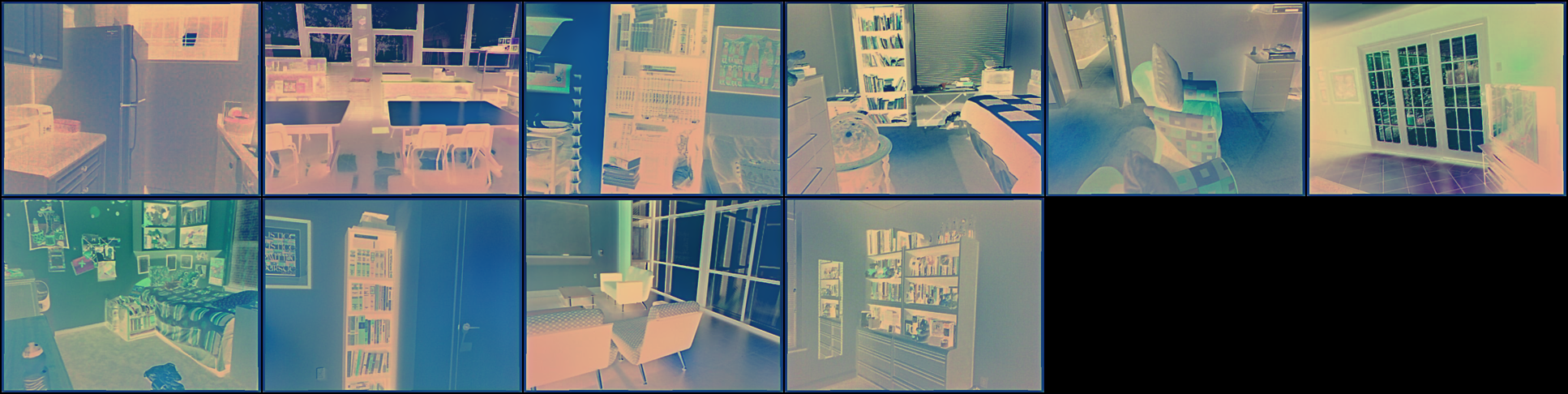} }
			\centering
			\vspace{-5mm}
			\caption{}
			\label{fig:technique_1_var_1_Residue_Pred_half}
		\end{subfigure}
		\begin{subfigure}[b]{\textwidth}
			\centering {\includegraphics[trim = 0.0cm 8.6cm 0.0cm 0.0cm, clip, scale = 0.18]{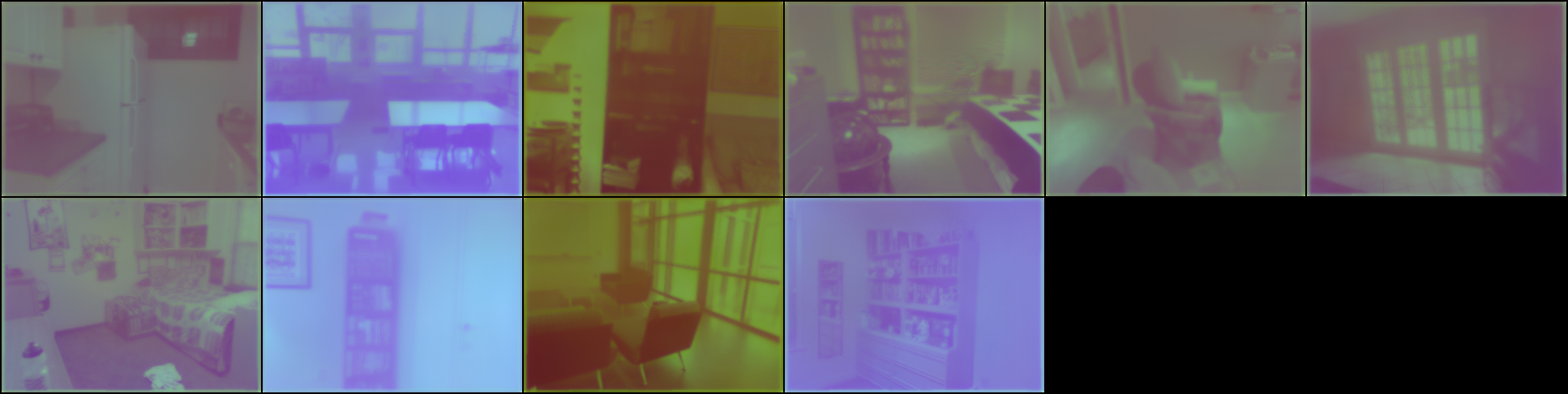} }
			\centering
			\vspace{-5mm}
			\caption{}
			\label{fig:technique_1_var_1_Complex_Pred_half}
		\end{subfigure}
		\caption[endProcess]{ \textbf{Qualitative Measures obtained by \textit{Proposed} Technique 1 Variant 1 :} (a) Original \emph{RGB} images (i.e. $J_c(\mathbf{x})$) (b) Ground Truth of ``Initial Degraded Image'' (i.e. $I_c(\mathbf{x})$) (c)  Ground Truth of ``Simulated Underwater Image'' (d) Predicted ``Initial Degraded Image'' (i.e. $I^{Degraded}_{Initial}$) (e) Predicted ``Residual Image'' (i.e. $I^{Residue}$) (f) Predicted ``Simulated Underwater Image'' (i.e. $\hat{I}^{Simulated}_{Predicted}$).  }
		\label{fig:all_single_task_1}
		\vspace{-0.5cm}
	\end{figure*}

\end{document}